\documentclass[preprint,5p,times]{elsarticle}
\biboptions{sort&compress}

\usepackage{balance}

\usepackage{xparse}
\usepackage{microtype}
\microtypesetup{
  stretch=30,
  shrink=30,
  step=3}
\usepackage{hyphenat}
\usepackage[british]{babel}
\usepackage[T1]{fontenc}

\usepackage{relsize}
\usepackage[scaled=.85]{zi4}
\usepackage{pifont}

\usepackage{amsmath,bm}
\usepackage{amssymb}
\usepackage{xspace}

\newcommand{\cmark}{\checkmark}
\newcommand{\xmark}{\ding{55}}

\usepackage{array}
\usepackage{booktabs}
\usepackage[table]{xcolor}

\usepackage{graphicx}
\usepackage{wrapfig}
\makeatletter
\def\fps@figure{t}
\makeatother

\usepackage[
  role=cardinal,
  byauthor
  ]{credits}
\usepackage[unicode,
  colorlinks=true,
  allcolors=black,
  urlcolor=blue
  ]{hyperref}
\hypersetup{
  pdftitle  = {ROS-related Robotic Systems Development with V-model-based Application of MeROS Metamodel},
  pdfauthor = {Tomasz Winiarski, Jan Kaniuka, Daniel Giełdowski, Jakub Ostrysz, Krystian Radlak, Dmytro Kushnir}
}
\usepackage{hypcap}

\NewDocumentCommand{\taginline}{m}{\hypertarget{#1}{\textbf{[#1]}}\xspace}
\NewDocumentCommand{\tagref}{m}{\hyperlink{#1}{[#1]}\xspace}

\newcommand{\Fig}[1]{Fig.~\ref{#1}}
\newcommand{\Tab}[1]{Tab.~\ref{#1}}
\newcommand{\Sec}[1]{Sec.~\ref{#1}}

\newcommand{\solution}[1]{\textsc{#1}}
\newcommand{\model}[1]{\textsf{#1}}
\newcommand{\diagram}[1]{{\textsf{\fontsize{9.5pt}{11pt}\selectfont #1}}}
\newcommand{\rosNode}[1]{\textsf{#1}}

\newcommand{\rosExePathName}[1]{\rosNode{#1}}
\newcommand{\rosNamespace}[1]{\rosNode{#1}}
\newcommand{\rosMessage}[1]{\textit{#1}}
\newcommand{\activity}[1]{\textit{#1}}
\newcommand{\actref}[2]{\textsf{[#1]} \textit{#2}}

\input{MeROS_VModel_HeROS-RAS-Stereotypes}

\journal{Robotics and Autonomous Systems}

\begin{document}

\hypersetup{
  linkcolor=black,
  citecolor=black,
  urlcolor=blue
}
\let\WriteBookmarks\relax

\begin{frontmatter}

\title{ROS-related Robotic Systems Development with V-model-based Application of MeROS Metamodel}

	\author[1]{Tomasz Winiarski\corref{cor1}}
	\ead{tomasz.winiarski@pw.edu.pl}
	\ead[url]{https://www.robotyka.ia.pw.edu.pl/team/twiniarski}
    \credit{Tomasz Winiarski}
    {Conceptualization, Methodology, Supervision, Resources, Project administration, Writing -- original draft, Funding acquisition, Writing -- review \& editing}

	\author[3]{Jan Kaniuka}
	\ead{jan.kaniuka@piap.lukasiewicz.gov.pl}
    \credit{Jan Kaniuka}
    {Software, Conceptualization, Formal analysis, Writing -- original draft, Validation, Visualization}

	\author[1]{Daniel Giełdowski}
	\ead{daniel.gieldowski@pw.edu.pl}
	\ead[url]{https://www.robotyka.ia.pw.edu.pl/team/dgieldowski}
    \credit{Daniel Giełdowski}
    {Software, Conceptualization, Formal analysis, Writing -- original draft, Validation, Visualization}

	\author[1]{Jakub Ostrysz}
  \ead{kuba.ostrysz@gmail.com}
    \credit{Jakub Ostrysz}
    {Software, Conceptualization, Formal analysis, Writing -- original draft, Validation, Visualization}

	\author[2]{Krystian Radlak}
  \ead{krystian.radlak@pw.edu.pl}
    \credit{Krystian Radlak}
    {Investigation, Writing -- original draft}

	\author[4]{Dmytro Kushnir}
	\ead{kushnir_d@ucu.edu.ua}
	\ead[url]{https://apps.ucu.edu.ua/en/teachers/dmytro-kushnir-2/}
    \credit{Dmytro Kushnir}
    {Validation, Methodology, Formal analysis, Investigation,  Writing -- original draft, Writing -- review \& editing}

	\affiliation[1]{organization={Warsaw University of Technology, Institute of Control and Computation Engineering},
		addressline={Nowowiejska 15/19},
		city={Warsaw},
		postcode={00-665},
		country={Poland}}

	\affiliation[2]{organization={Warsaw University of Technology, Institute of Computer Science},
		addressline={Nowowiejska 15/19},
		postcode={00-665},
		city={Warsaw},
		country={Poland}}

	\affiliation[3]{organization={Industrial Research Institute for Automation and Measurements PIAP},
		addressline={Aleje Jerozolimskie 202},
		city={Warsaw},
		postcode={02-486},
		country={Poland}}

	\affiliation[4]{organization={Ukrainian Catholic University, Applied Science Faculty},
		addressline={Kozelnytska 2a},
		city={Lviv},
		postcode={79026},
		country={Ukraine}}

  \cortext[cor1]{Corresponding author. \textit{Email:} \texttt{tomasz.winiarski@pw.edu.pl} (T. Winiarski)}

\begin{abstract}
  Systems built on the Robot Operating System (ROS) are increasingly easy to assemble, yet hard to govern and reliably coordinate. Beyond the sheer number of subsystems involved, the difficulty stems from their diversity and interaction depth. In this paper, we use a compact heterogeneous robotic system (HeROS), combining mobile and manipulation capabilities, as a demonstration vehicle under dynamically changing tasks. Notably, all its subsystems are powered by ROS.
  The use of compatible interfaces and other ROS integration capabilities simplifies the construction of such systems. However, this only addresses part of the complexity: the semantic coherence and structural traceability are even more important for precise coordination and call for deliberate engineering methods. The Model-Based Systems Engineering (MBSE) discipline, which emerged from the experience of complexity management in large-scale engineering domains, offers the methodological foundations needed.
  Despite their strengths in complementary aspects of robotics systems engineering, the lack of a unified approach to integrate ROS and MBSE hinders the full potential of these tools. Motivated by the anticipated impact of such a synergy in robotics practice, we propose a structured methodology based on MeROS---a SysML metamodel created specifically to put the ROS-based systems into the focus of the MBSE workflow. As its methodological backbone, we adapt the well-known V-model to this context, illustrating how complex robotic systems can be designed with traceability and validation capabilities embedded into their lifecycle using practices familiar to engineering teams.
\end{abstract}

\begin{graphicalabstract}
  \includegraphics[width=\textwidth]{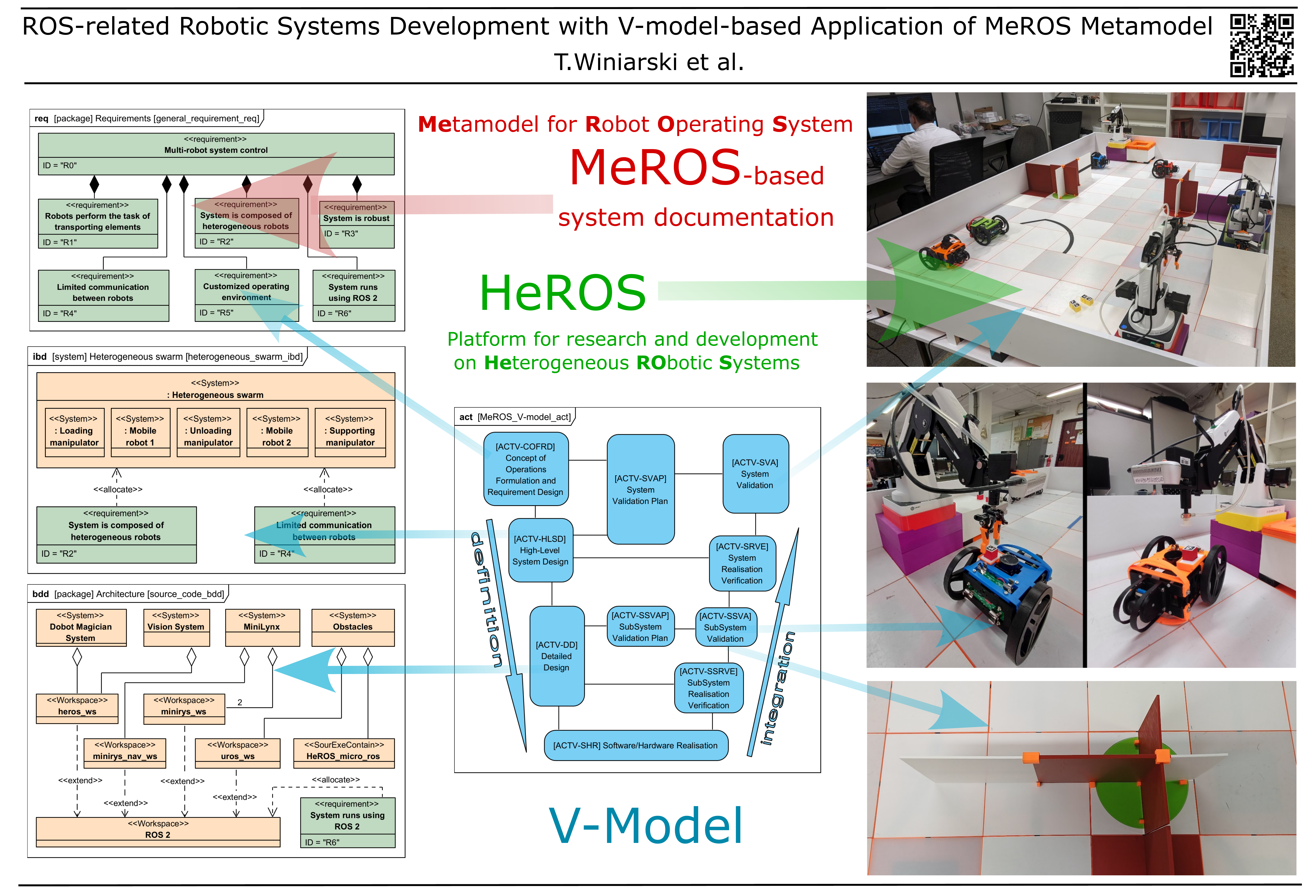}
\end{graphicalabstract}


\end{frontmatter}

\section{Introduction}
\label{sec:intro}

  The evolution of engineering methods has always been designed not only to boost system performance but also to streamline how systems are designed, verified, and maintained.
  As robotics matures from early experimental platforms to real-world, often safety-critical deployments, the need for structured engineering practices has become pressing. Modern robotic systems interlace the software, embedded hardware, sensors, and actuator components together, often replicating those relations across both simulation and physical domains. While component-level tools in robotics have matured, the methodologies for managing system-level integration and lifecycle traceability remain fragmented \cite{Brugali2015MdseInRobotics,Casalaro2022ModelDriven}. Addressing these challenges is largely left to individual teams striving for consistent and reusable design patterns across their projects.

  Several middleware and software integration frameworks have been developed to manage robotic system complexity. The Robot Operating System (\solution{ROS}) \cite{Quigley2009ROS}, with its successor \solution{ROS~2}~\cite{Macenski2022ROS2}, are the most widely adopted among them. \solution{ROS} exemplifies how modular design, open standards, and collaborative tooling can shape an entire field. With ecosystems that combine loose coordination with active maintenance driven by research labs, individual contributors, and major industry stakeholders, it has become the de facto standard runtime environment. As Brooks and Kaupp observed, ``Standards become powerful when a critical mass of users arises, such that people choose to conform due to the advantages that it brings.''~\cite{BrooksKaupp2005} \solution{ROS} reached that mass by providing modular integration and component reuse, but it did so without enforcing formal modelling or lifecycle discipline. That social and technical momentum gives ROS great practical value, but it also reveals limits: \solution{ROS} and its ecosystem were not originally designed to enforce formal lifecycle discipline, model-based traceability, or system-level verification---gaps that complicate development as applications become safety-critical and requirements evolve.

  The Systems Engineering (SE) discipline emerged in large-scale aerospace and defence contexts in the mid-20th century, offering a unifying methodology for addressing these challenges~\cite{INCOSE2023Handbook}. Yet, adapting SE to the needs of ROS-based robotics agile workflows and hybrid system behaviour remains methodologically unresolved and missing key components. The V-model~\cite{INCOSE2023Handbook} is one of the most recognised lifecycle structures in SE. It links system definition steps with matching stages of integration and validation. While it is well known in safety-critical domains, its core principles apply more broadly: iterative design refinement, traceability, and structured development. Many fields have adopted or proposed iterative forms of the V-model to balance engineering discipline with development agility \cite{Graessler2020VDI2206}.

  Even though \solution{ROS}-based robotic systems development increasingly employs complex lifecycle features like co-simulation and digital twin integration, approaches that multiply interlinked representations still lack the essential frameworks to keep those replicas in sync and align system-level models with executable architectures. This disconnect complicates verification, traceability, and solutions support across development cycles. This paper builds on the Model-Based Systems Engineering (MBSE) foundation using a model-based approach tailored to \solution{ROS} development and builds on prior works in robotics and cyber-physical systems, including tools like EARL \cite{Winiarski2020EARL} and SPSysML~\cite{Dudek2025SpSysML}.

  This work introduces a development method structured around a \solution{MeROS}-tailored V-model. Proposed methodology extends the existing modelling language with the guidance of structured system design phases, while not strictly imposing any rigid procedure, leaving its definition to the developers of the particular system. Referencing previously framed in MBSE literature the three ``pillars'' that hold it: (1) an integrated \textit{language}, (2) modelling \textit{methodology}, and (3) \textit{tooling} \cite{Delligatti2013SysML}. While the metamodel for ROS (\solution{MeROS})~\footnote{\url{https://github.com/twiniars/meros}}~\cite{Winiarski2023MeROS} already delivers a SysML-based language and leverages standard toolchains, it previously lacked a formalised methodology. This contribution addresses that gap by tailoring a missing method to the settled-in-advance modelling language and tool, thus completing the triad. It is among the first methodologies that connect a \solution{ROS} Platform-Specific Metamodel (PSM) with a generalised V-model application framework, establishing a foundation for traceable system design while remaining flexible to different workflow contexts.

  \paragraph{Structure of the article}
  \Sec{sec:v-model} presents the rationale and structure of the \solution{MeROS}-tailored V-model. The following two sections provide a practical illustration of the application of this V-model in \solution{ROS}-related systems documented with \solution{MeROS}. \Sec{sec:heros} describes the \model{HeROS} testbed, which serves as the ground-truth for practically validating our methodology. Then, in \Sec{sec:application}, we systematically and step-by-step demonstrate the feasibility and end-to-end traceability of the robotics system development process. \Sec{sec:related-work} traces the historical origins of various development approaches for robotics systems, situating our work within the evolution of the field. \Sec{sec:discussion} directly compares our methodology with previous approaches, including a summary table and critical analysis of strengths and limitations. Finally, \Sec{sec:conclusions} summarises the paper.

\section{MeROS-tailored V-model}
\label{sec:v-model}

\begin{figure}
	\centering
	\includegraphics[scale = 0.7]{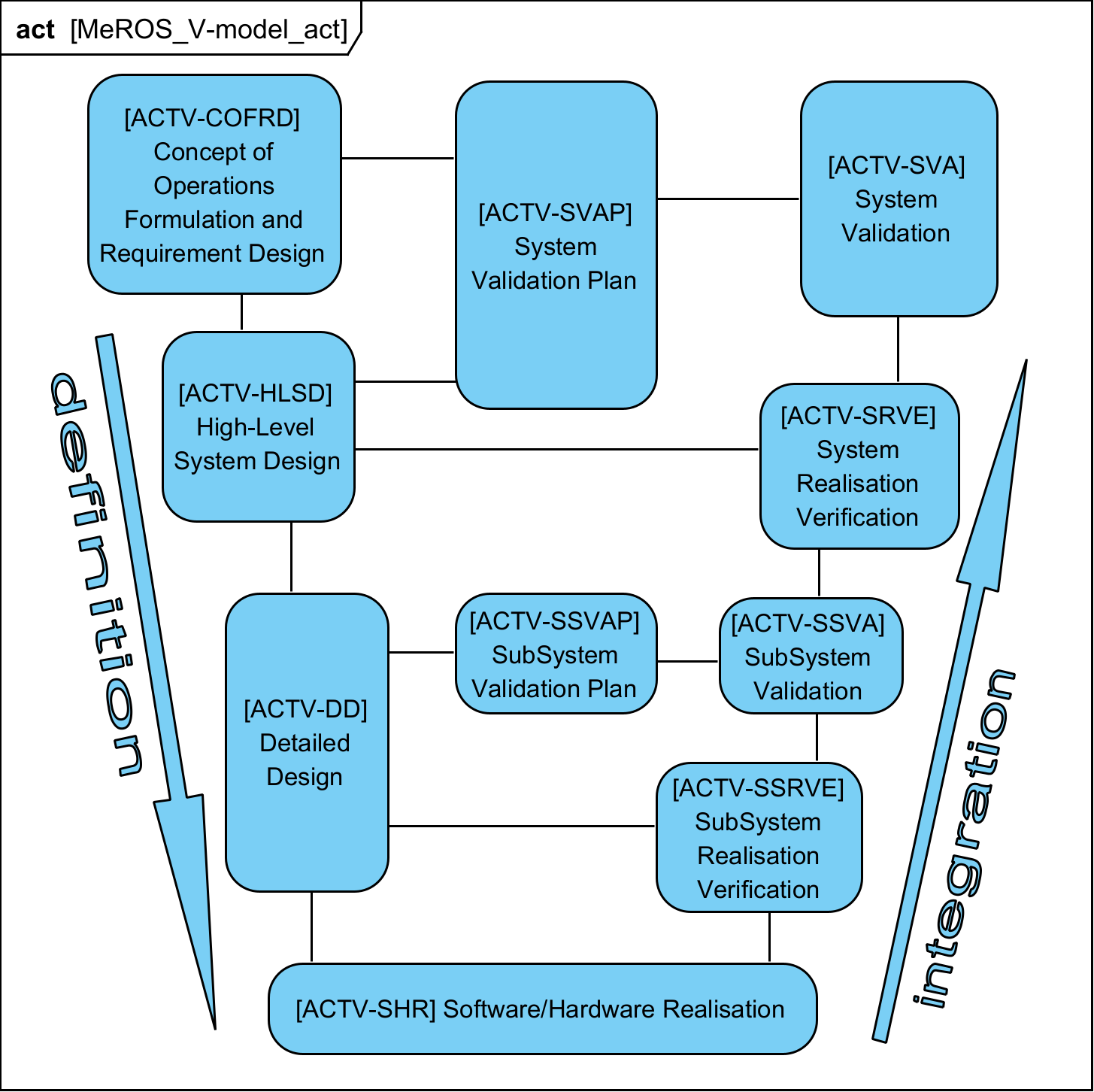}
	\caption{MeROS-tailored V-model}
	\label{fig:MeROS-V-model}
\end{figure}

The \solution{MeROS}-tailored V-model (\Fig{fig:MeROS-V-model}) provides a development methodology designed explicitly for \solution{ROS}-based Systems. By adapting classical V-model principles to the needs of robotics, it integrates SysML-based modelling for traceable and reusable System design, ensuring a consistent link between high-level requirements and detailed implementation within the \solution{MeROS} ecosystem. The model itself was developed according to a set of keys for usability and methodological coherence, denoted as [RV1]--[RV4] below:

\begin{itemize}
\itemsep -0.1em
    \item \mbox{[RV1]} --- addressing diverse interpretations of the V-model, particularly in terms of how verification~\&~validation are understood;
    \item \mbox{[RV2]} --- simplicity and generality;
    \item \mbox{[RV3]} --- supporting flexible integration into diverse pro\-ject work\-flows, without enforcing a specific procedural path;
    \item \mbox{[RV4]} --- possibility of direct application with \solution{MeROS}.
\end{itemize}

The requirement [RV1] addresses the broad range of V-model variants found in systems engineering literature. Particularly, it reflects differing views on verification---as either a stepwise transition between design stages~\cite{Cederbladh2024Early} or a test of conformance between design and implementation---and on validation as the assessment of a realised system against functional and operational expectations~\citep{Zheng2016Interface}.

To guide the reader across both textual narrative and diagrams, we use semantically tagged elements such as \stSystem{} and \stNode{} consistently throughout the paper, when referring to formal \solution{MeROS} metamodel elements.

The order in which the actions from \Fig{fig:MeROS-V-model} are invoked crucially depends on the specific project and the assumed design procedure \cite{INCOSE2023Handbook}. Hence [RV3], in the description below, we focus on actions rather than sequences of transitions.

\subsection{System definition with validation plan}

This set of actions relates to the definition of the System (\stSystem{}) at various levels, together with the plan for its validation. For this purpose, the SysML and UML diagrams are used, particularly those created in compliance with the \solution{MeROS} metamodel.

\begin{itemize}

	\item \taginline{ACTV-COFRD} --- The stage of Concept of Operations formulation and requirement (\stRequirement{}) design is the level at which, in principle, elements defined directly in MeROS are not used. The specification primarily employs:
   (i) \diagram{requirements diagrams} and
   \diagram{behavioural diagrams}, mainly:
   (ii) \diagram{use case diagrams},
   (iii) \diagram{activity diagrams} and
   (iv) \diagram{sequence diagrams}.
  These are the basis for both the formulation of the validation plan, \tagref{ACTV-SVAP} and the overall design of the \stSystem{}, \tagref{ACTV-HLSD}, and its individual parts, \tagref{ACTV-DD}.

  It should be noted that, in the subsequent phases of the \stSystem{} definition, each \stRequirement{} formulated in this stage must be \stAllocate{} to the relevant \stSystem{} elements responsible for its fulfilment (example in \Fig{fig:heterogeneous_swarm_ibd}).

  \item \taginline{ACTV-HLSD} --- In this stage, the general view of the \stSystem{} specification is presented, consistent with the Concept of Operations and \stRequirement{} that were formulated in \tagref{ACTV-COFRD}. Here, the \solution{MeROS}-related elements are introduced. The main \stSystem{} is defined in a general way, i.e., especially its internal \stSystem{} elements and Communication Channels (\stCommChannel{}) are proposed. For that purpose, all the SysML diagrams are helpful, with emphasis on:
      (i) \diagram{block definition diagrams},
      (ii) \diagram{internal block diagrams},
      (iii) \diagram{activity diagrams},
      (iv) \diagram{sequence diagrams}, and
      (v) \diagram{use case diagrams}.

  \item  \taginline{ACTV-SVAP} --- The plan of the whole \stSystem{} validation. Its specification can be supported by \diagram{behavioural diagrams}, especially \diagram{sequence diagrams}. The plan can be specified in terms of the System Concept of Operations from the \tagref{ACTV-COFRD} and \tagref{ACTV-HLSD} stages.

  \item \taginline{ACTV-DD} --- Here, the general view of the \stSystem{} formulated in \tagref{ACTV-HLSD} is decomposed and specified with more detailed elements. For example, the structure of high-level \stSystem[Sub] elements and generic \stCommChannel[][s] is modelled using specific \solution{ROS} communication interfaces: Topics (\stTopic[][s]), Services (\stService[][s]), and Actions (\stAction[][s]). Similarly, the internal behaviour of the \stSystem{} can be described at the level of ROS Nodes (\stNode[][s]) and their interactions.

  \item \taginline{ACTV-SSVAP} --- The validation plan for the particular \stSystem[Sub][s] and their parts' validation was created in a way analogous to \tagref{ACTV-SVAP}, according to \tagref{ACTV-DD}.

\end{itemize}

\subsection{System realisation with verification and validation}

The actions below relate to system realisation. It should be noted that this encompasses not only physical implementations but also simulation-based models (often referred to as digital twins when integrated with real-time data).

\begin{itemize}
    \item \taginline{ACTV-SHR} --- Realisation (implementation) of the \stSystem{} design, including both software and hardware.
    \item \taginline{ACTV-SSRVE} --- Verification of the architecture (structure and behaviour) of particular \stSystem{} elements and their parts.
    \item \taginline{ACTV-SSVA} --- Validation of particular \stSystem{} elements and their functionalities according to the plan specified in \tagref{ACTV-SSVAP}.
    \item \taginline{ACTV-SRVE} --- Verification of the architecture (structure and behaviour) of the entire \stSystem{}.
    \item \taginline{ACTV-SVA} --- Validation of the entire \stSystem{} functionality according to the plan specified in \tagref{ACTV-SVAP}.
\end{itemize}

\section{HeROS platform}
\label{sec:heros}

\begin{figure}
	\centering
	\includegraphics[width=\columnwidth]{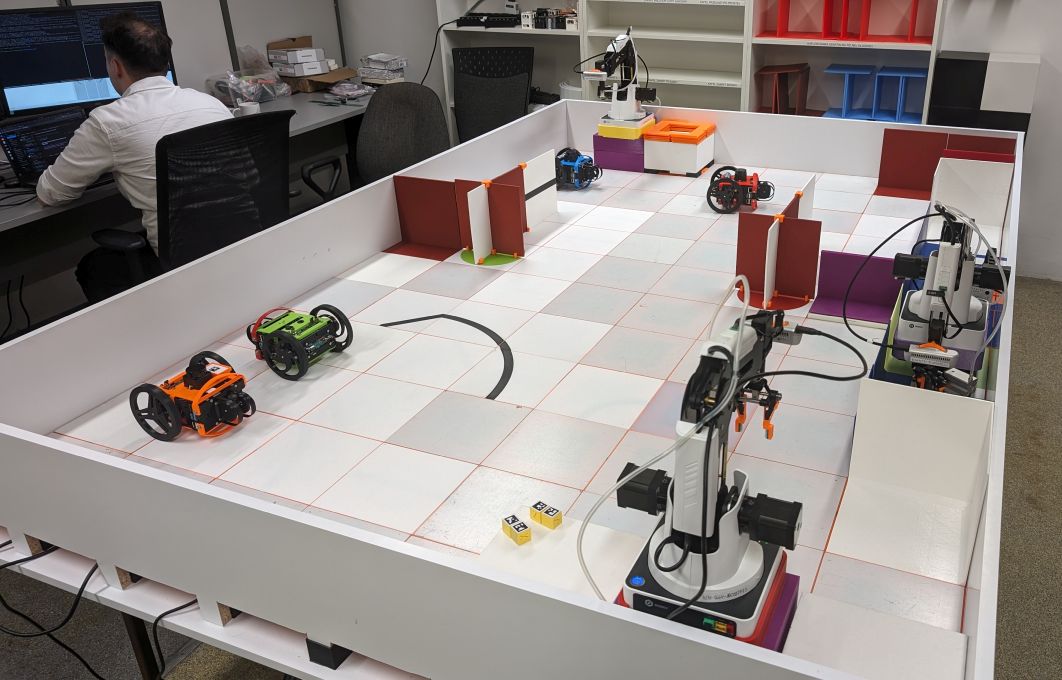}
	\caption{Configuration of the HeROS board and robots employed in the experiments presented in this article.}
	\label{fig:heros-board}
\end{figure}

The \solution{HeROS} \stSystem{} is a miniaturised, low-cost physical test platform for heterogeneous robotic systems \cite{Winiarski20204HeROS} (\Fig{fig:heros-board}), which was created as a small-scale robotic testbed \citep{Xu2025Testbeds} with the specific features:

\begin{itemize}
	\itemsep -0.1em
		\item modular, tile-based board that facilitates the easy reproduction of experiments in various environmental configurations,
		\item tiles small enough to construct complex test environments within standard building spaces,
		\item easy access to the interior of the board from all sides,
		\item moderate board price with small-scale manufacturing,
		\item simple to produce and modify,
		\item possibility to route cables underneath the tiles,
		\item mobile robots that match the dimensions of the board tiles,
		\item manipulation robots that are compatible with the dimensions of the board tiles,
		\item unified software framework for all of the hardware.
	\end{itemize}

\subsection{Board with tiles}
\label{subsec:board_with_tiles}

 The board is constructed using 3D-printed legs, bases, and tiles. The environment is surrounded by wooden walls. In its current form, the board's legs are 10 cm high. Thanks to this, it is possible to place any vulnerable components required to power the robots and utilise the environment in a secure and isolated space. Each base is supported by four legs and fits one square-shaped, 20 cm-wide tile. Currently available tiles include flat floor (with different markings for purposes of various robotic applications) and tiles with 20 cm-high walls, turns or bends. This allows for building complex testing environments in a finite space. Furthermore, the mechanical sliding rail, utilised together with the \model{Dobot Magician} manipulator (\Sec{subsec:customises_dobot}), was also integrated into the board. Currently, the board can be used for designing test environments of an area up to \(160\,\mathrm{cm} \times 240\,\mathrm{cm}\).

\subsection{Mechanised obstacles}
\label{subsec:mechanised_obstacles}

\begin{figure}
	\centering
	\includegraphics[width=\columnwidth]{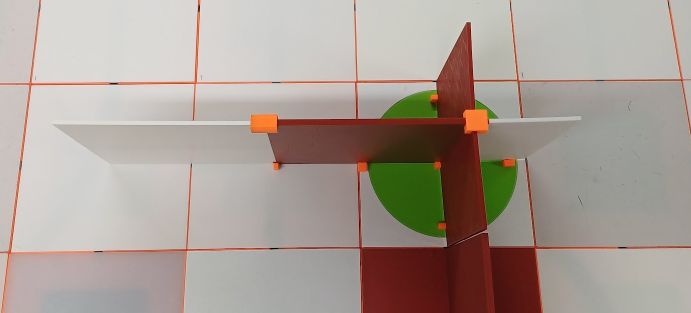}
	\caption{Rotating obstacle (top view).}
	\label{fig:rot_obstacle}
\end{figure}

Moving obstacles can be placed on the board to dynamically reconfigure the operating environment. Moving obstacles are constructed using 3D-printed walls and a rotating platter. An example of a moving obstacle, designed in the shape of the letter “T” and positioned horizontally on the plate, is shown in \Fig{fig:rot_obstacle}. It is actuated by a DC motor with a gear, mounted in a dedicated socket beneath the plate. In the experiments considered in this article, there are two such obstacles on the board, together forming a double-leaf gate in the middle of the board. Obstacle control is implemented using the \model{ESP32} microcontroller and the micro-ROS \cite{Belsare2023Micro_ROS} framework.

\subsection{Customised Dobot Magician robot}
\label{subsec:customises_dobot}

\begin{figure}
	\centering
	\includegraphics[scale=0.15]{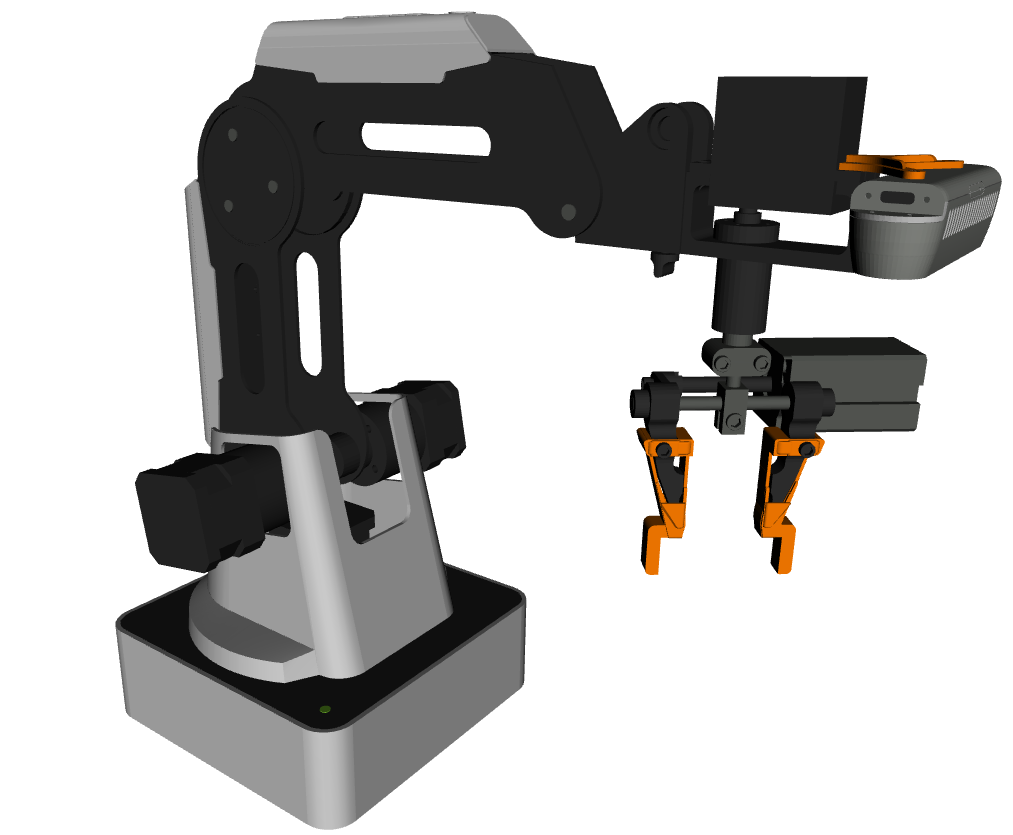}
	\caption{Customised \model{Dobot Magician} robot---visualisation~\cite{Winiarski20204HeROS}.}
	\label{fig:dobot}
\end{figure}

The \model{Dobot Magician} is a 4-DOF serial educational manipulator (\Fig{fig:dobot}). This off-the-shelf platform supports various end-effectors, including a two-finger gripper, a pneumatic suction cup, and a soft gripper. Our manipulators were enhanced with custom 3D-printed accessories, such as gripper extenders, a mount for the Intel RealSense D435i depth camera, and fixtures for signal cables and compressor tubing---all integrated into the custom \solution{ROS~2}-based control system and movement planner. Furthermore, a \solution{ROS~2}-based control system with a graphical user interface was implemented\footnote{\url{https://vimeo.com/793400746}} to expand the robot's functionality \cite{Kaniuka2023BSc}. The kinematic configuration of the \model{Dobot Magician} makes it particularly well-suited for standard Pick-and-Place operations, i.e., straightforward point-to-point object manipulation tasks.

\subsection{MiniLynx mobile robot}

\begin{figure}
	\centering
	\includegraphics[scale=0.25]{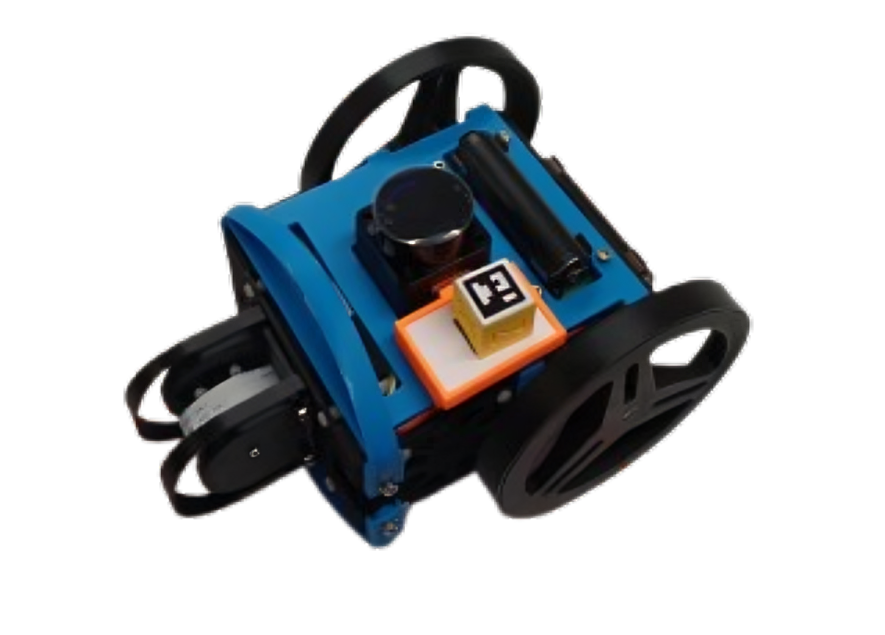}
	\caption{\model{MiniLynx} (MiniRyś) robot.}
	\label{fig:minirys}
\end{figure}

The \model{MiniLynx}\footnote{\url{https://vimeo.com/1052300450}} (\model{MiniRyś} in Polish) is a small differential drive mobile robot (\Fig{fig:minirys}) designed for research and development of multi-robot systems. It has the ability to drive in two modes of locomotion: a vertical mode, where the robot balances on two wheels, and a horizontal mode, where the third point of support is one of the bumpers. The robot is characterised by its ability to detect obstacles, map its surroundings using the SLAM algorithm, navigate autonomously, and collaborate with other units to perform swarm navigation tasks. It also includes a variety of sensors such as LiDAR, an RGB camera, and an IMU, and its computing unit is a Raspberry Pi~4B single-board computer. The robot's software architecture is realised with \solution{ROS~2} components spanning from the device-level locomotion control and sensor processing through to the high-level mission coordination, encompassing both navigation and behaviour selection.

\section{Application}
\label{sec:application}

The concept of the V-model makes it possible to systematically organise the development process of a multi-robot system. It facilitates the connection between the system definition and integration and testing stages. In the following, the development process of a multi-robot heterogeneous system was chosen as a representative example.

The purpose of the following description is to highlight representative actions from the proposed V-model, rather than covering the whole project in detail.

\subsection{Concept of operations formulation and requirements design \tagref{ACTV-COFRD}}

\begin{figure}
	\centering
	\includegraphics[scale=0.22]{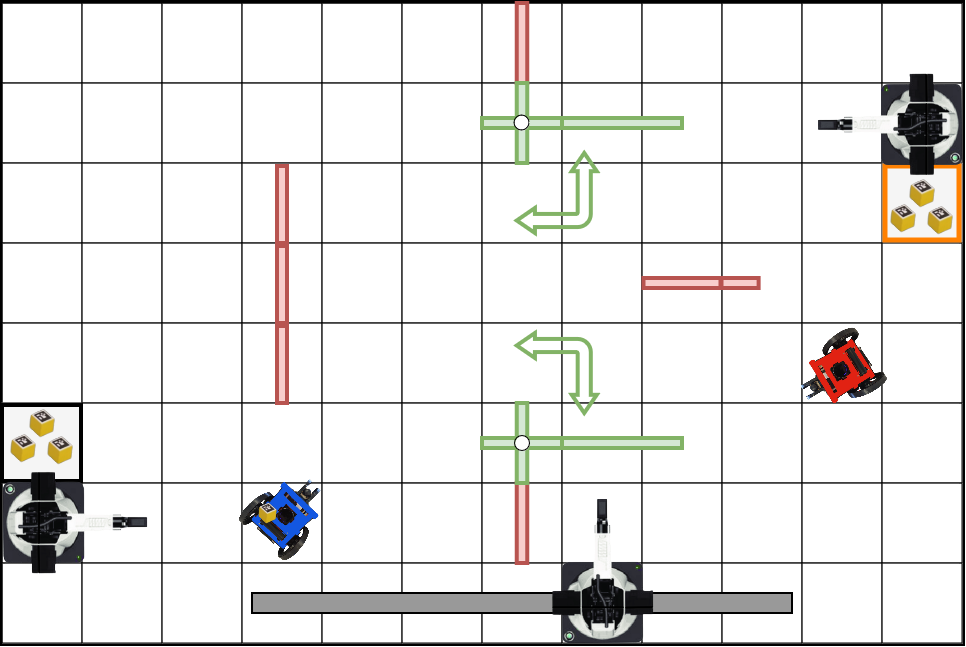}
	\caption{Operating environment layout (top view).}
	\label{fig:board_sketch}
\end{figure}

This case study was developed as a demonstrator for the proposed methodology and a validation scenario for the \model{HeROS} research platform. While synthetic in setup, the scenario emulates real-world multi-agent logistics tasks in constrained and dynamic environments, providing a representative testbed for structured modelling and validation using MeROS and the V-model.
The mission involved coordinated cooperation between heterogeneous robots: two mobile \model{MiniLynx} robots and three \model{Dobot Magician} manipulators. Robots shall cooperatively carry out the task of transporting cubes from one end of a board to the opposite. The \stSystem{} should continue to operate despite a failure of one \model{Mobile robot} and adapt to changes in the operational environment. The operational board layout is shown in \Fig{fig:board_sketch}.

\begin{figure}
	\centering
	\includegraphics[scale=0.65]{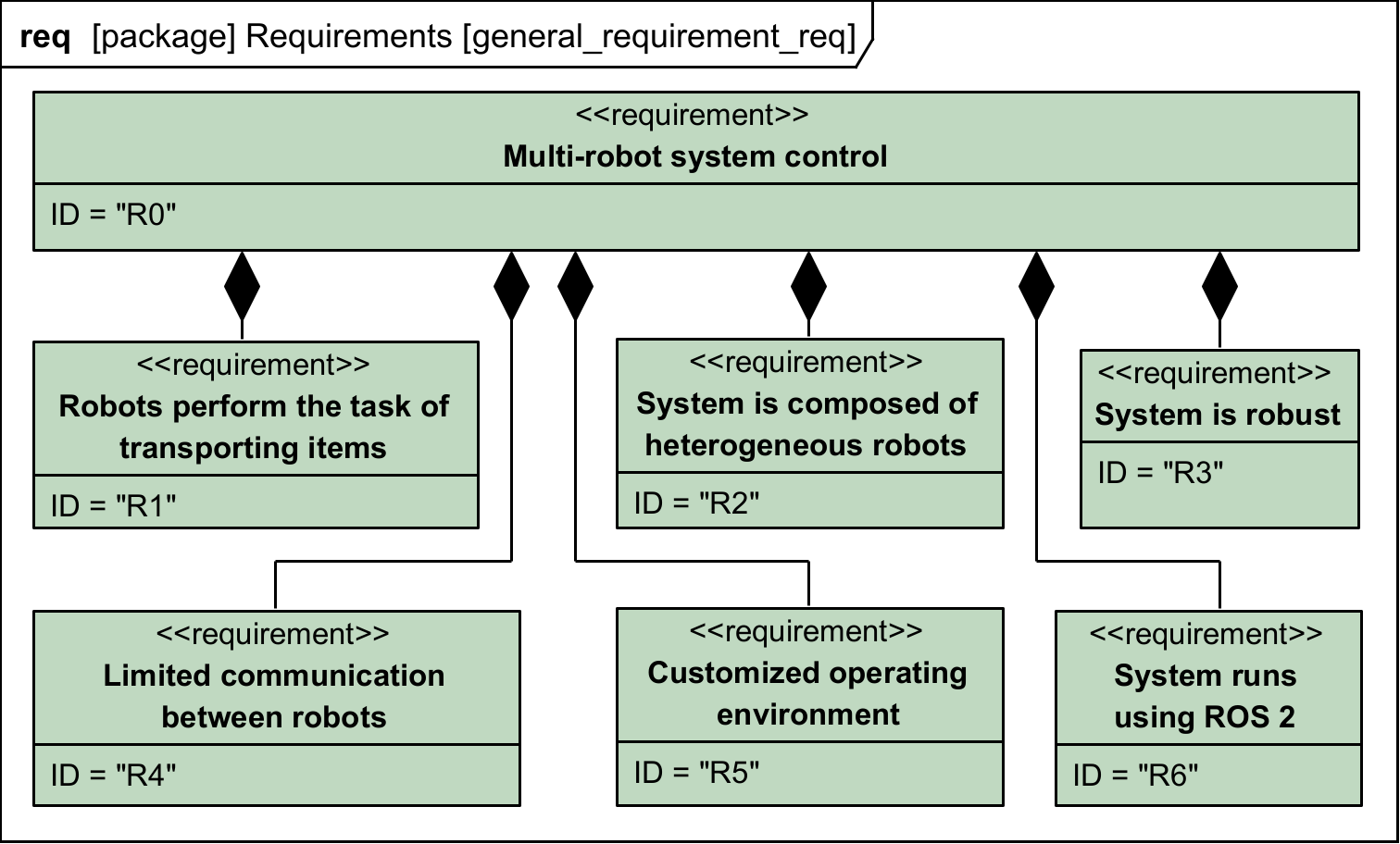}
	\caption{General \stRequirement{} set for the designed \stSystem{}.}
	\label{fig:general_req}
\end{figure}

\Fig{fig:general_req} presents the general \stRequirement[][s] set for the designed \stSystem{}. Among them, \stRequirement{} \model{[R1]} specifies the operational goal: \model{Robots perform the task of transporting items}.
 The primary \stRequirement{} is \model{[R2]: \stSystem{} compositions of heterogeneous robots}, which are able to operate under \model{Limited communication between the robots [R4]}. In addition, the created \model{System should be robust [R3]} and capable of operating in a \model{Customised operating environment [R5]}. Regarding the software side, the \model{System should be based on ROS~2 [R6]}.

\subsection{High-level system design \tagref{ACTV-HLSD}}

After formulating the \stRequirement{}, the next step is to describe the operation of the \stSystem{} at the highest level of abstraction; at this stage, design details are intentionally omitted. The operation of the \stSystem{} can be represented using \diagram{behavioural diagrams}, in this case, are \diagram{activity diagrams}.

To describe individual behaviours, we first need to identify all the actors and components that should perform those behaviours and are going to be involved in interactions within the overall \stSystem{}.

\begin{figure}
	\centering
	\includegraphics[scale=0.65]{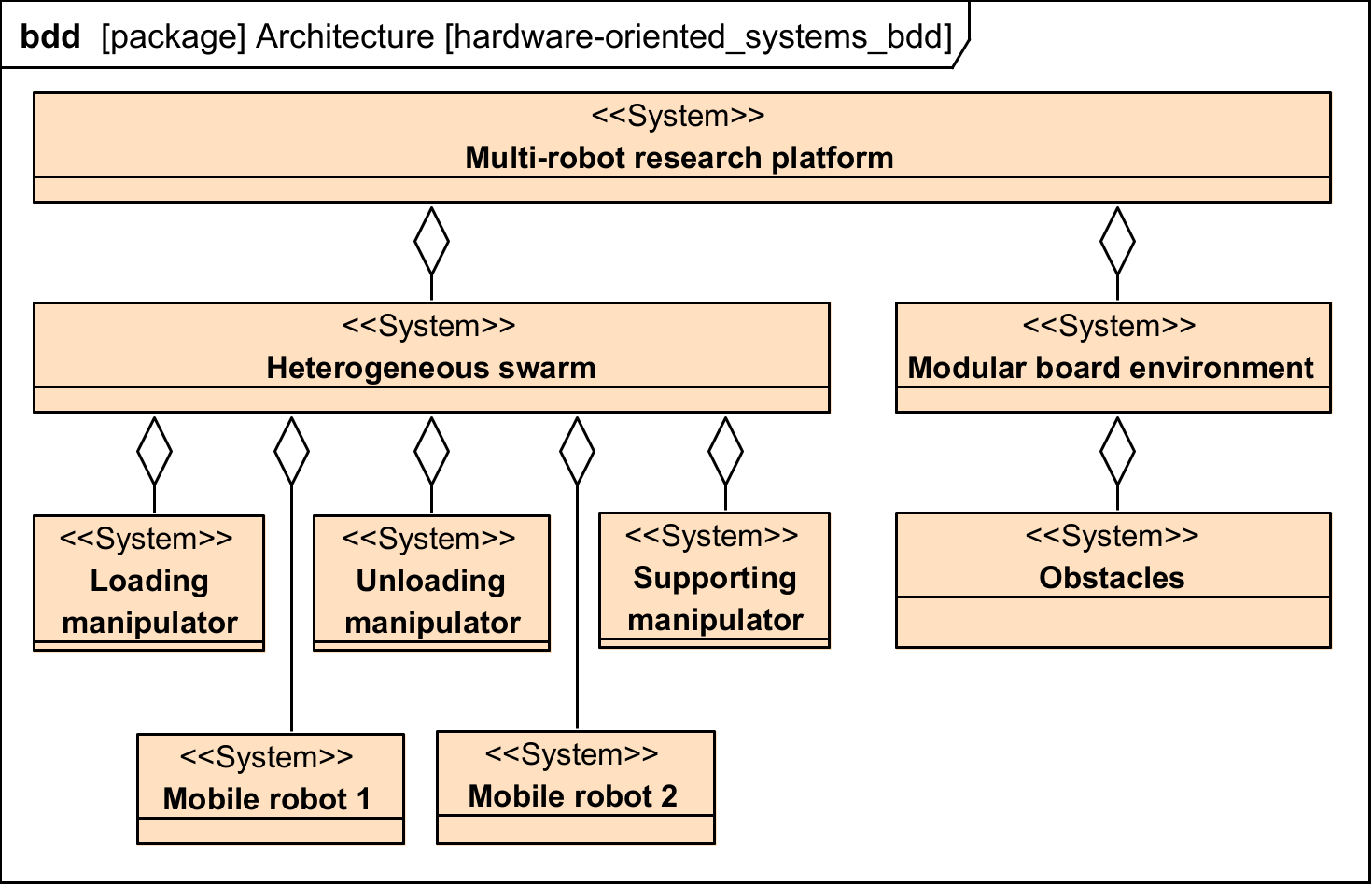}
	\caption{\stHardware{}-oriented structure of the top-level \stSystem{}.}
	\label{fig:hardware-oriented_systems_bdd}
\end{figure}

\Fig{fig:hardware-oriented_systems_bdd} describes the~\model{Multi-robot research platform} \stSystem{} structure that references two \stSystem[][s]: the~\model{Heterogeneous swarm} and the~\model{Modular board environment}. The \model{Heterogeneous swarm}, in turn, groups together three \model{Manipulator}s and two \model{Mobile robot} \stSystem[][s]. While the \model{Modular board environment} coordinates only one \model{Obstacles} \stSystem{}.

\begin{figure}
	\centering
	\includegraphics[scale=0.65]{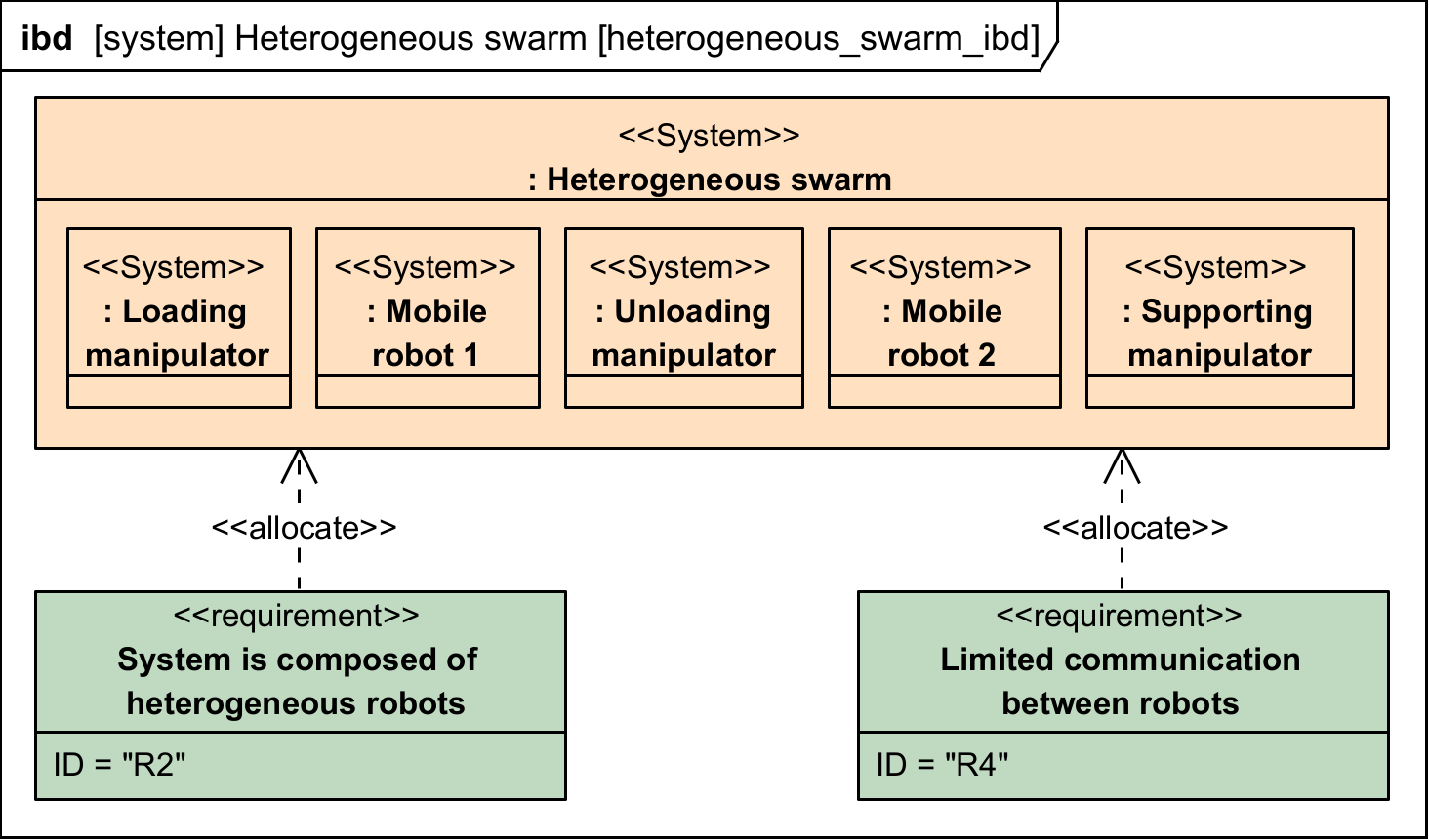}
	\caption{\diagram{Internal block diagram} of the~\model{Heterogeneous swarm} \stSystem.}
	\label{fig:heterogeneous_swarm_ibd}
\end{figure}

There is no explicit peer-to-peer communication among the \model{Heterogeneous swarm} \stSystem{} parts; therefore, there is no \stCommChannel{} shown in the \Fig{fig:heterogeneous_swarm_ibd}. All the coordination emerges implicitly, without a centralised controller. Instead, the physical cube serves as a medium for communication, as long as its presence can be detected by both \model{Mobile robots} and \model{Manipulators}.

\begin{figure}
	\centering
	\includegraphics[scale=0.65]{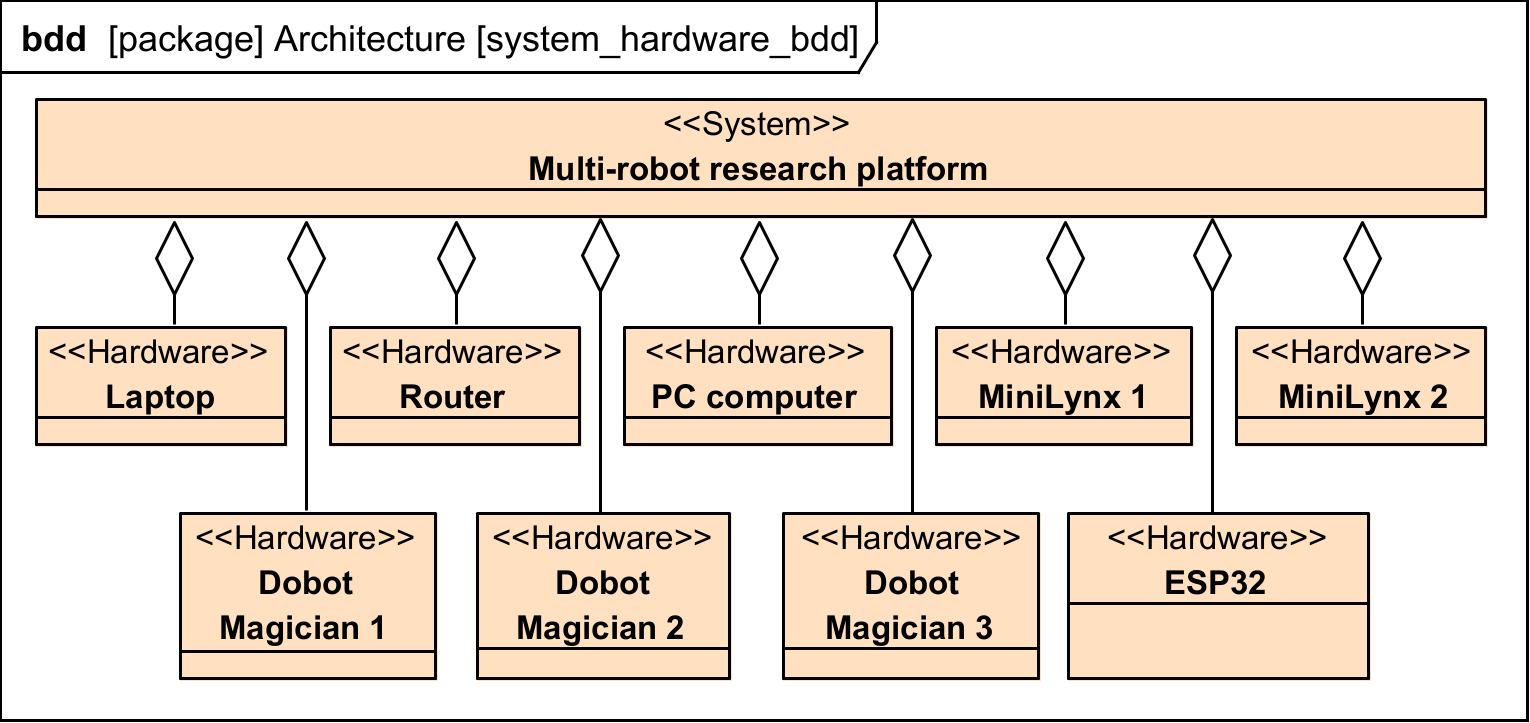}
	\caption{\stHardware{} components of the \model{Multi-robot research platform}.}
	\label{fig:system_hardware_bdd}
\end{figure}

\begin{figure}
	\centering
	\includegraphics[scale=0.65]{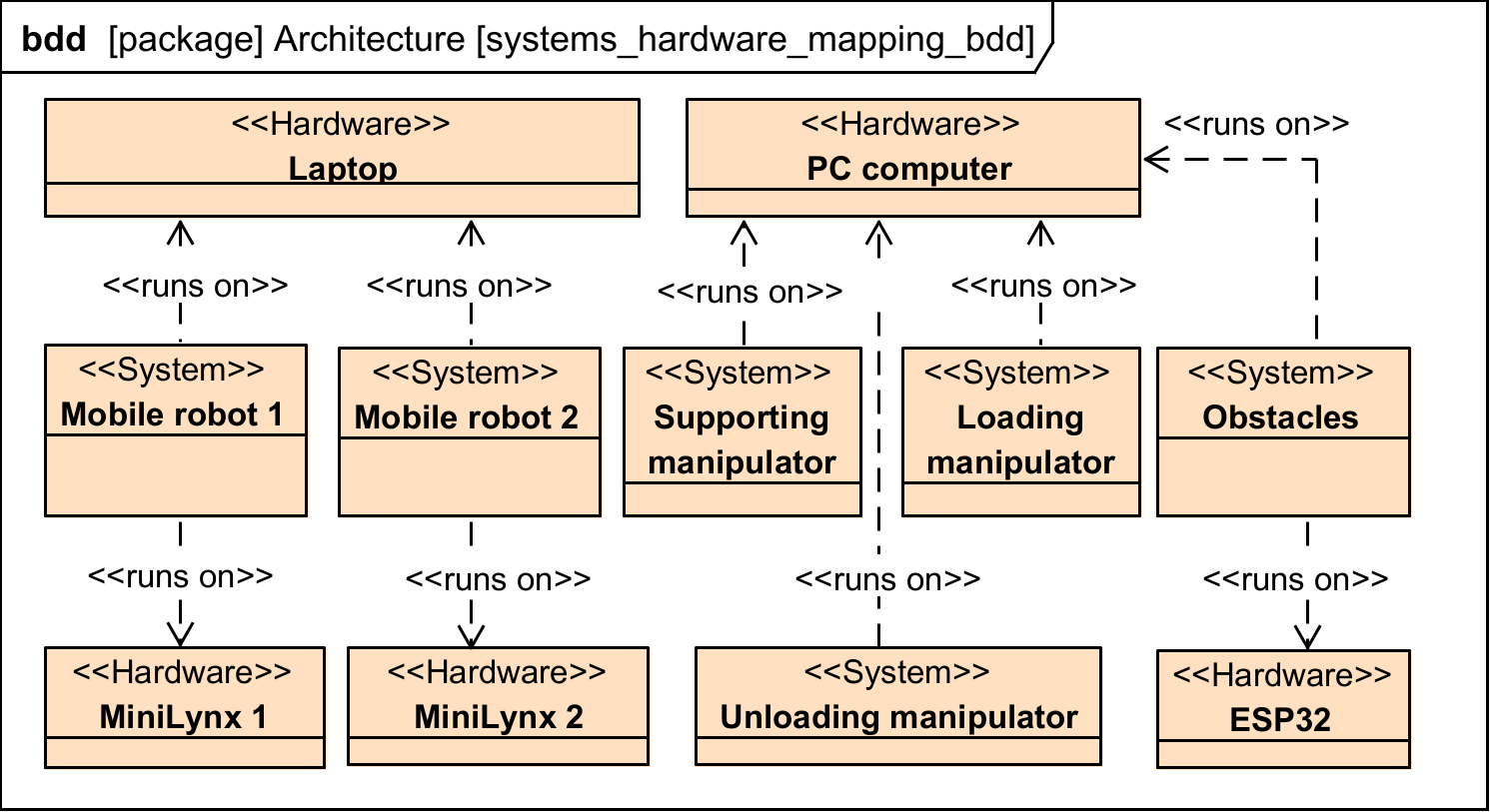}
	\caption{\stSystem{}-\stHardware{} mapping.}
	\label{fig:systems_hardware_mapping_bdd}
\end{figure}

Before discussing the behavioural aspects of all our \stSystem[][s], we need to explore the \stHardware{} layer. The \Fig{fig:system_hardware_bdd} depicts \stHardware{} network configuration at the system level. Then \Fig{fig:systems_hardware_mapping_bdd} further details on which parts of the \stSystem{} are hosted by which \stHardware{} elements, tracing the correspondence between \stSystem{}-level functions and their realisation as \stHardware{}.

\begin{figure}
	\centering
	\includegraphics[scale=0.65]{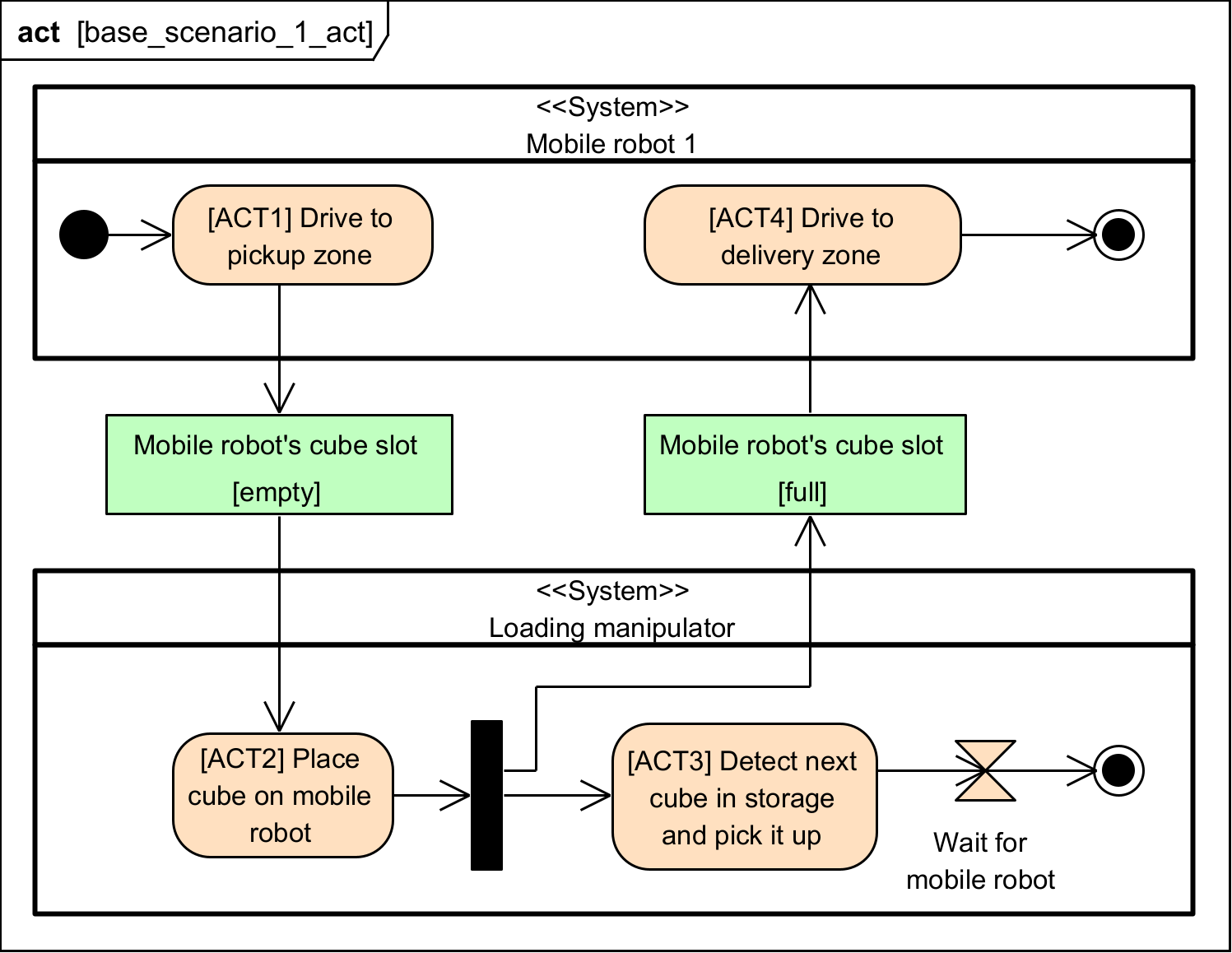}
	\caption{General operation of the \model{Heterogeneous swarm} \stSystem{} (part~1).}
	\label{fig:base_scenario_1_act}
\end{figure}

\begin{figure}
	\centering
	\includegraphics[scale=0.65]{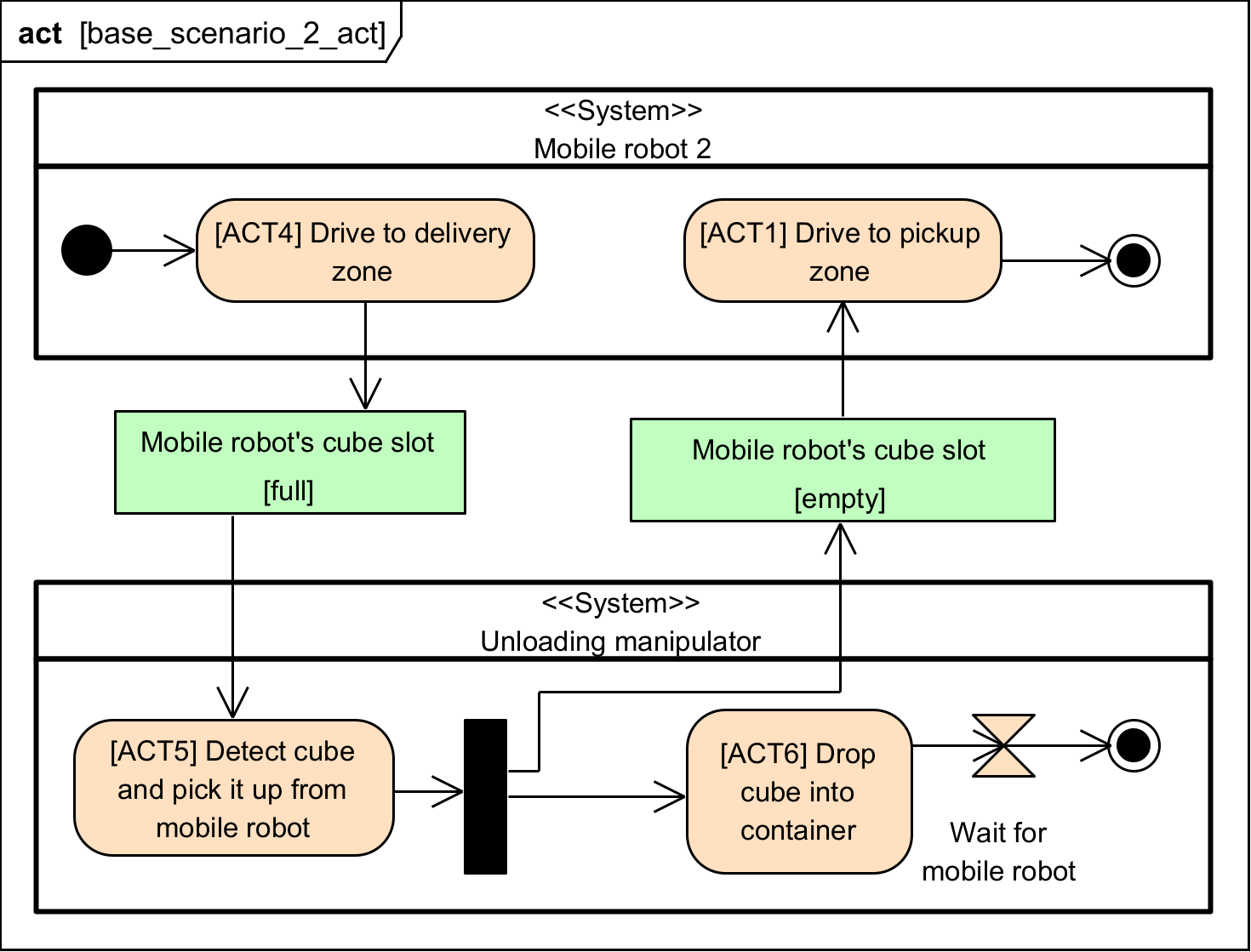}
	\caption{General operation of the \stSystem{} (part 2).}
	\label{fig:base_scenario_2_act}
\end{figure}

As the illustrative Use Case was chosen the task of collaborative cube transportation involved coordinating the efforts of the \model{Heterogeneous swarm}. The goal is to organise robust cube delivery by \model{Mobile robots} between the \model{Manipulator} units. The flow of this behaviour is divided into two \diagram{activity diagrams}, as shown in \Fig{fig:base_scenario_1_act} and \Fig{fig:base_scenario_2_act}.

In the initial state, the \model{Loading manipulator} holds a~cube in its gripper and is positioned such that its camera can observe the environment: the specific section of the board. At the same time, the \model{Mobile robot} is \actref{ACT1}{Driving to the pickup zone}, while the \model{Loading manipulator} is awaiting the \model{Mobile robot} to come within its operational range.

The \model{Mobile robot} does not explicitly notify about loading readiness; instead, the \model{Loading manipulator} infers this based on its camera data. When the \model{Mobile robot} is within~the \model{Manipulator}'s range, the \model{Manipulator} \actref{ACT2}{Places a~cube onto the mobile robot}, and subsequently \actref{ACT3}{Detects the next cube in storage and picks it up}. The \model{Mobile robot}, based on the readings from the LiDAR, recognises that a~cube is on it and then \actref{ACT4}{Drives to the delivery zone} towards the \model{Unloading manipulator}.

At the same time, on the opposite side of the board, the \model{Unloading manipulator} and \model{Mobile robot 2} handle the unloading operation.
\model{Mobile robot 2} \actref{ACT4}{Drives autonomously to the delivery zone}, where the \model{Unloading manipulator} observes the board with its camera.

When the \model{Mobile robot} is in range of the \model{Unloading manipulator}, \actref{ACT5}{Detects the cube and picks it up from Mobile robot} and then \actref{ACT6}{Drop the cube into the container}. The \model{Mobile robot} will realise on its own that it is no longer carrying the cube and proceeds to \actref{ACT1}{Drive to the pickup zone} to begin the cycle again with the \model{Loading manipulator}.

\begin{figure}
	\centering
	\includegraphics[scale=0.65]{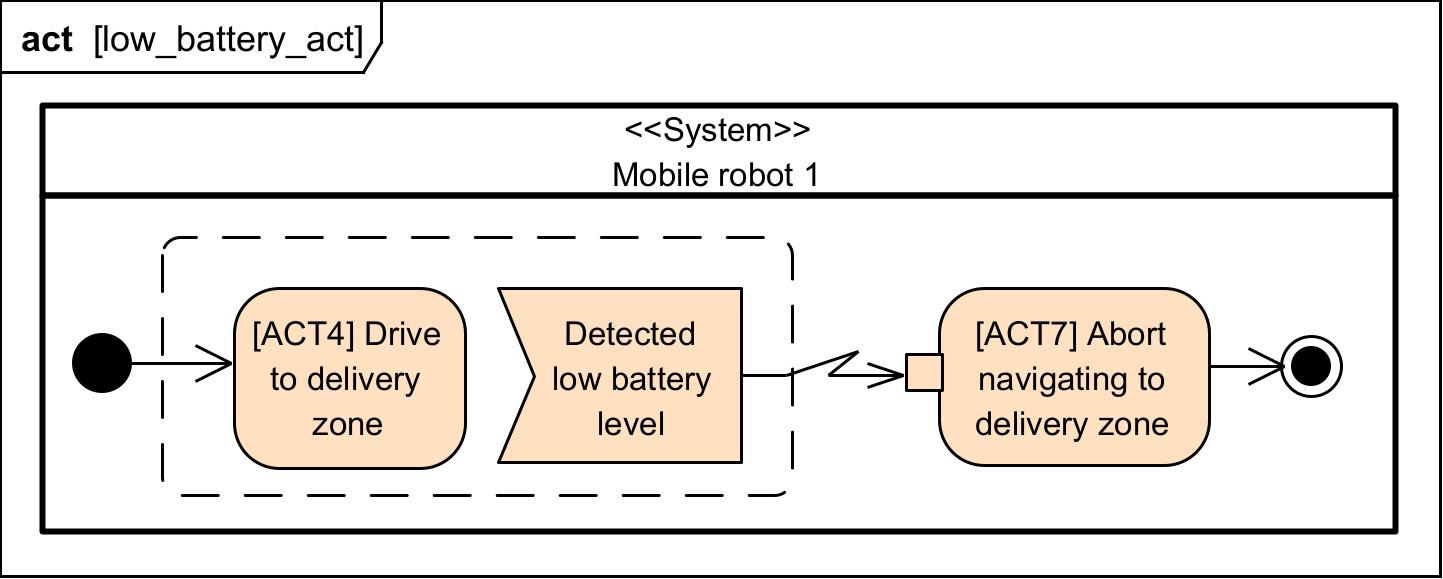}
	\caption{\model{Mobile robot} failure case (shown for \model{Mobile~robot~1}).}
	\label{fig:low_battery_act}
\end{figure}

The behaviour depicted in \Fig{fig:low_battery_act} describes an emergency situation in which a~\model{Mobile robot} may find itself while performing a~task, i.e., battery discharge. Each \model{Mobile robot} continuously monitors the voltage level of its battery. While driving to one of the goals (in this case, to the \model{Unloading manipulator} while \actref{ACT4}{Driving to the delivery zone}), if the battery level becomes too low to continue operation, the robot should instead \actref{ACT7}{Abort the navigation to the target}.

\begin{figure}
	\centering
	\includegraphics[scale=0.65]{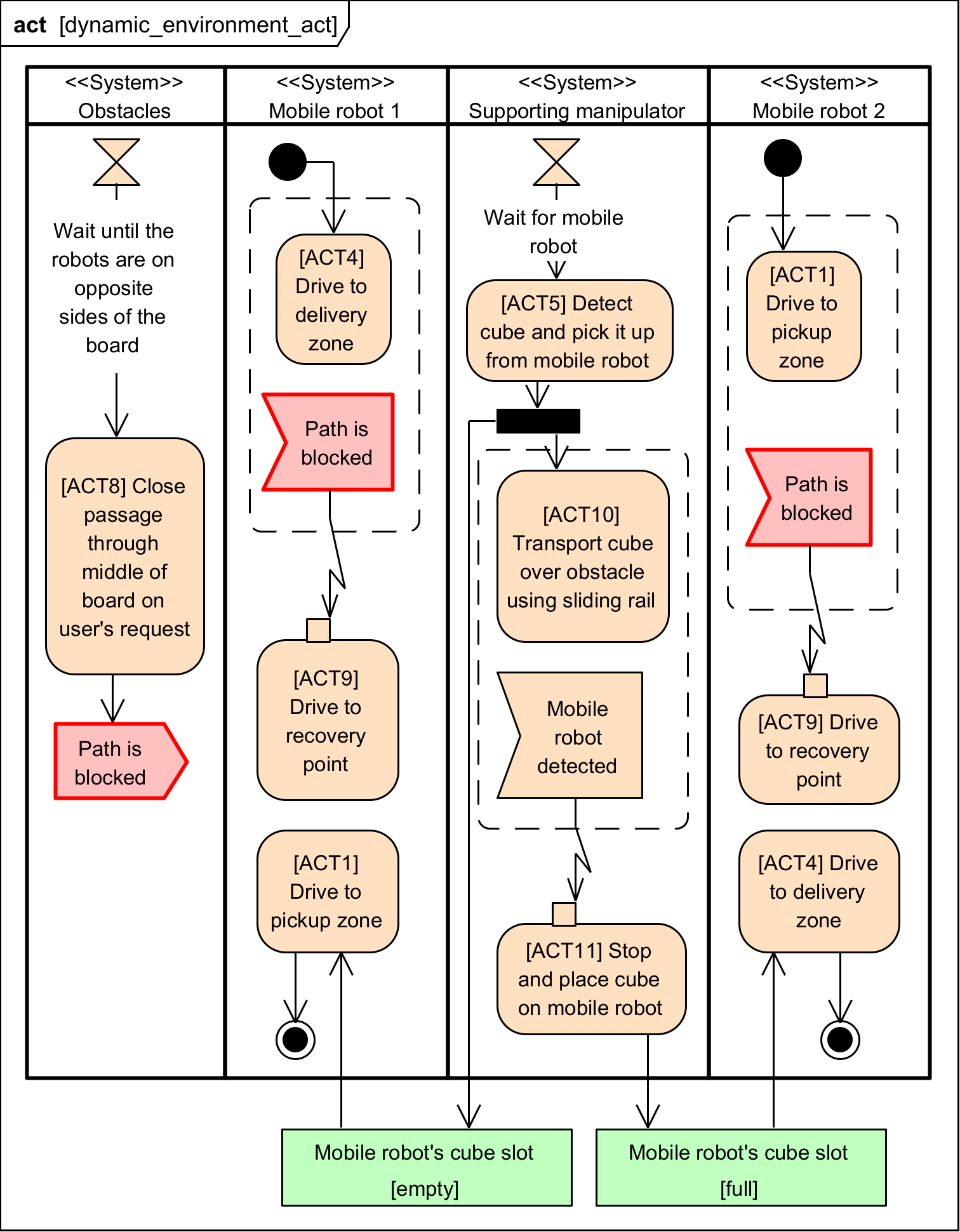}
	\caption{Occurrence of changes in~the operational environment.}
	\label{fig:dynamic_environment_act}
\end{figure}

The most extensive set of behaviours is shown in \Fig{fig:dynamic_environment_act}. It describes a~scenario in which the operational environment changes dynamically during the scenario execution. In such a~case, a~third Dobot Magician, mounted on the sliding rail, plays the role of \model{Supporting manipulator} to assist the other robots in performing their tasks. The set of behaviours depicted in the diagram begins as in the general operation scenario described earlier (see \Fig{fig:base_scenario_1_act} and \Fig{fig:base_scenario_2_act}). At some point, the system tester deliberately modifies the environment by triggering the environment change with the remote control. The \model{Obstacles} \stSystem{} \actref{ACT8}{Closes the passage through the middle of the board on the user's request}, dividing the test environment into two parts. The two \model{Mobile robots} cannot reach their target locations under these conditions. When the global planner module of the navigation subsystem fails to generate a new trajectory, it switches behaviour to \actref{ACT9}{Drive to the recovery point}.

Each \model{MiniLynx} robot has a~predefined recovery point stored in memory---an emergency location to drive to if the task cannot be completed \actref{ACT9}. These points lie within the operational range of the \model{Supporting manipulator}, which moves along a~sliding rail and continuously observes its surroundings via a~camera. When the \model{Mobile robot 1} encounters an \model{Obstacle} and cannot proceed to the delivery zone, it drives to its recovery point. Once there, the \model{Supporting manipulator} \actref{ACT5}{Detects the cube and picks it up from the Mobile robot}. It then lifts the cube as high as possible and \actref{ACT10}{Transports the cube over the obstacle using the sliding rail}.

At~the same time, the \model{Mobile robot 2}, from~which the cube has been removed, \actref{ACT1}{Drives to the pickup zone} to the \model{Loading manipulator} to transport another cube. Suppose it discovers that the \model{Path is blocked}, it \actref{ACT9}{Drives to its recovery point}. The \model{Supporting manipulator} awaits with a cube for the \model{Mobile robot 2}, and when it appears within~the range of the \model{Manipulator}, it will \actref{ACT11}{Place the cube on the Mobile robot}.

After getting the cube, \model{Mobile~robot~2} will \actref{ACT4}{Drive directly to the delivery zone}, where the \model{Unloading manipulator} will unload it, and the~\model{Supporting manipulator} on the sliding rail will return to its initial position (\textit{return act is not depicted on the diagrams}).

\subsection{Detailed design \tagref{ACTV-DD}}

\begin{figure}
	\centering
	\includegraphics[scale=0.65]{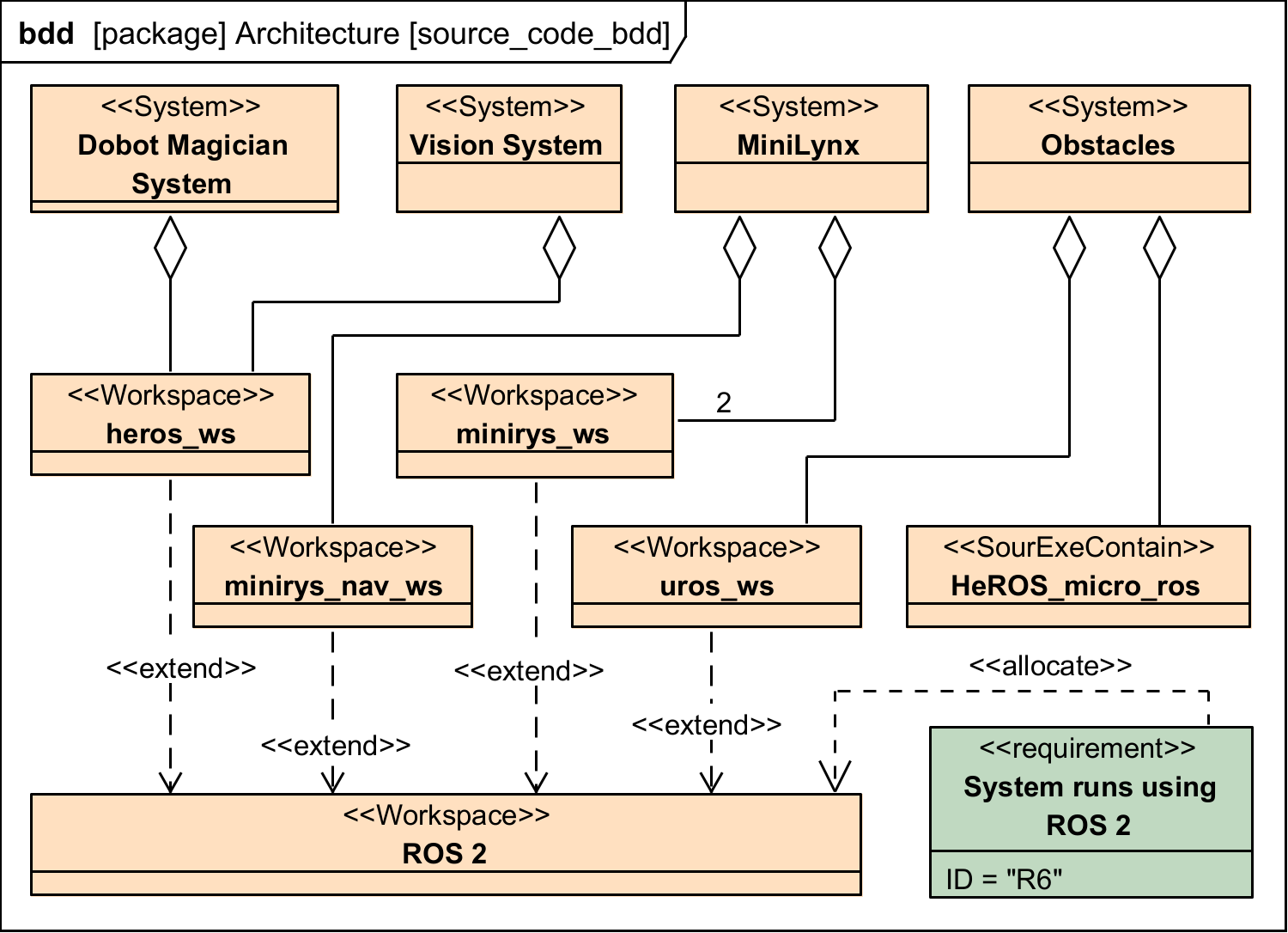}
    \caption{Source code divided between \stSrcExeContain{}.}
	\label{fig:source_code_bdd}
\end{figure}

\begin{figure}
	\centering
	\includegraphics[scale=0.65]{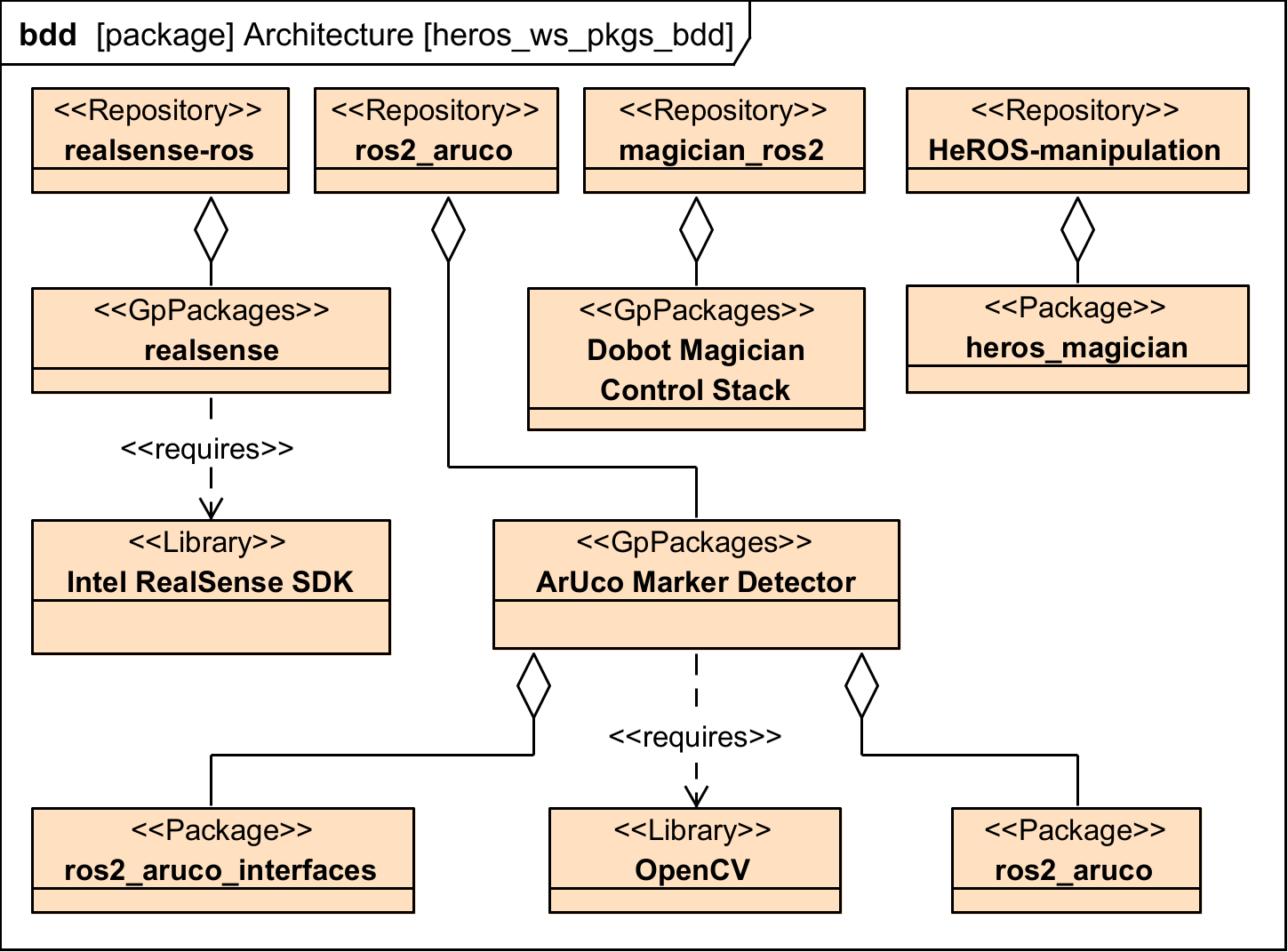}
	\caption{\solution{ROS~2} \stPackage{} inside the \rosExePathName{heros\_ws} \stWorkspace{}.}
	\label{fig:heros_ws_pkgs_bdd}
\end{figure}

The Detailed Design (\tagref{ACTV-DD}) stage focuses on structurally specifying the architecture of the \stSystem{}, elaborating on diagrams developed during the initial design (\tagref{ACTV-HLSD}) stage. This includes identifying \stSystem[Sub][s] within each \stSystem{} and defining the \stCommChannel[][s] that coordinate their interactions---preferably illustrated using \diagram{internal block diagrams}. The \solution{MeROS} metamodel facilitates this process by providing dedicated modelling constructs for both architectural and behavioural elements, as demonstrated in \Fig{fig:source_code_bdd} and \Fig{fig:heros_ws_pkgs_bdd}. At this stage, newly specified artefacts are grouped into appropriate Sources and Executable Containers (\stSrcExeContain{}): Packages (\stPackage{}), Repositories (\stRepository{}), and Work\-spaces (\stWorkspace{}). The fine-grained \stRequirement[][s] must be systematically allocated (\stAllocate{}) to concrete model elements to ensure traceability and support iterative verification across all architectural description layers.

\begin{figure}
	\centering
	\includegraphics[scale=0.65]{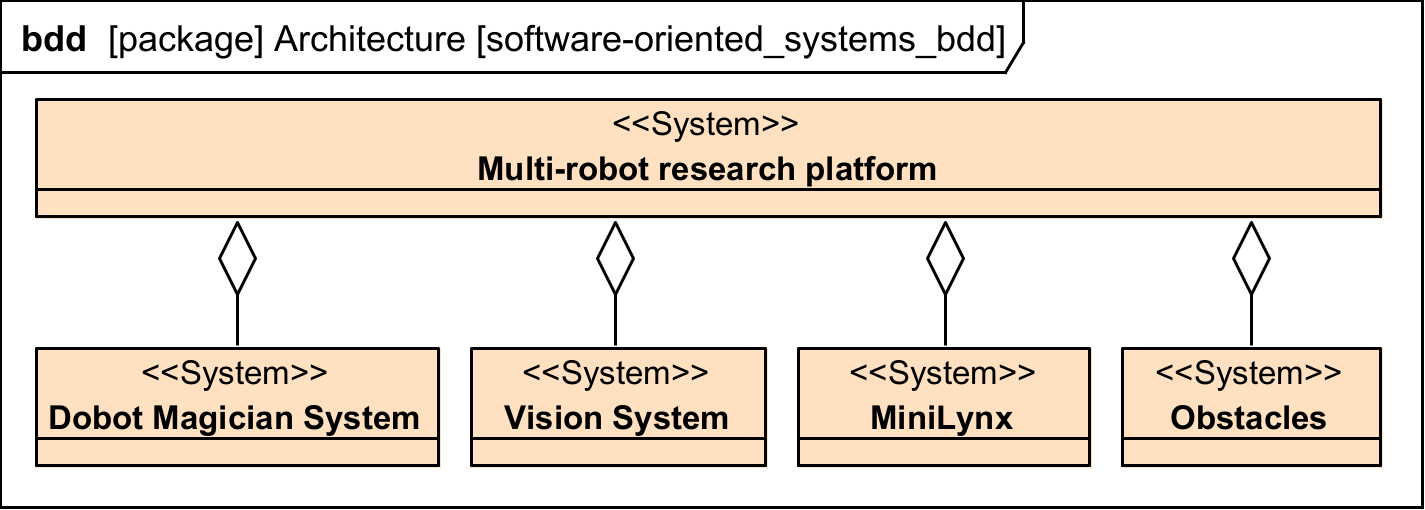}
	\caption{\stSystem{} composition oriented towards major software components.}
	\label{fig:software-oriented_systems_bdd}
\end{figure}

\Fig{fig:software-oriented_systems_bdd} presents the component structure of a~\model{Multi-robot research platform} \stSystem{}.
\model{Dobot Magician System} and \model{Vision System} are \stSystem[][s] associated with the \stHardware{} of the  \model{Dobot Magician manipulator}. Each \model{MiniLynx} \stSystem{} refers to \model{Mobile robot}, and the \model{Obstacles} \stSystem{} refers to the moving \model{Obstacles} controlled by the \model{Modular board environment}.

\begin{figure}
	\centering
	\includegraphics[scale=0.6]{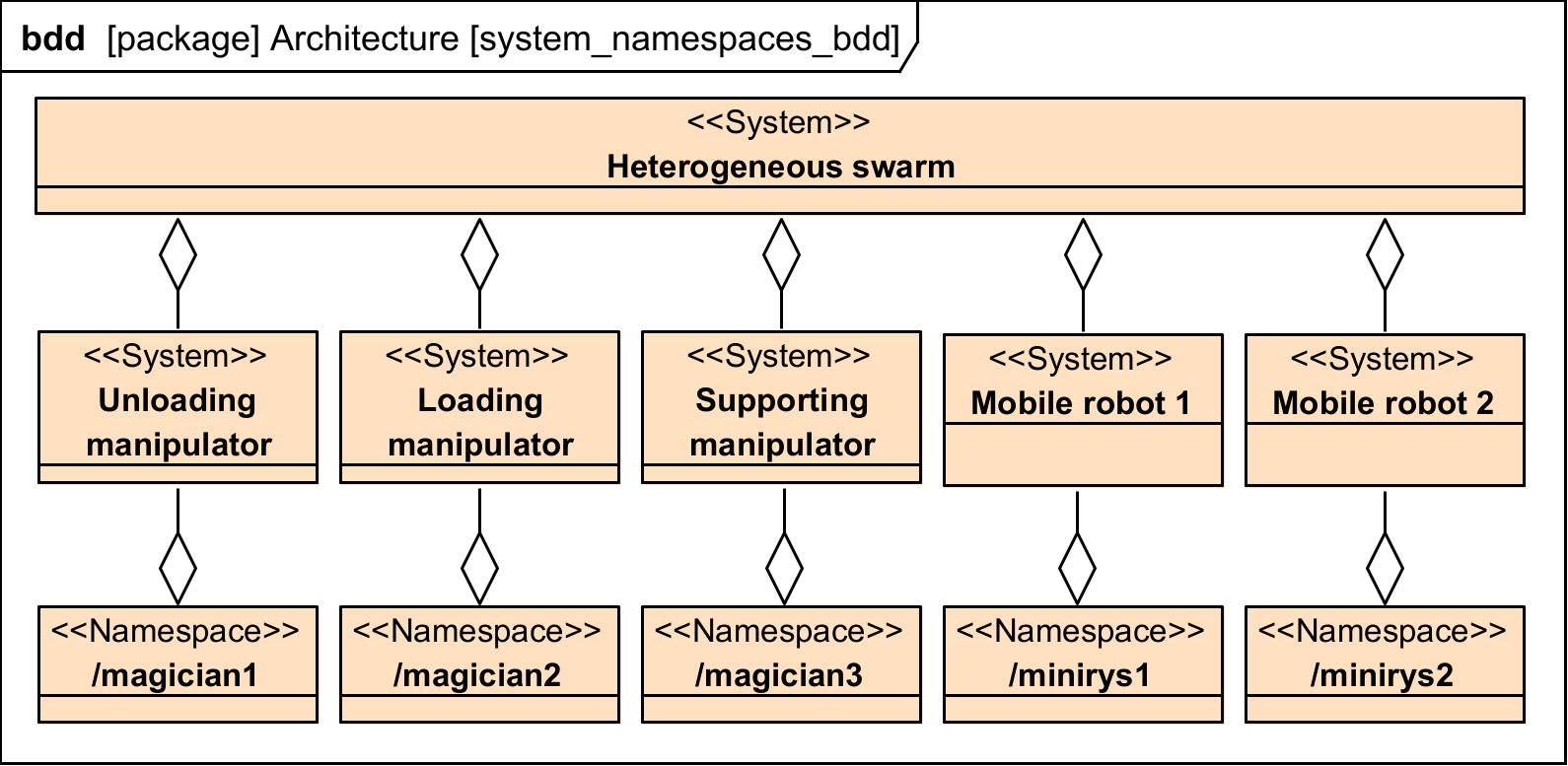}
	\caption{\stSystem[][s] with the assigned \stNamespace[][s].}
	\label{fig:system_namespaces_bdd}
\end{figure}

The individual \model{Mobile} and \model{Manipulator} robot \stSystem[][s] are deployed in \stNamespace[][s] as shown in \Fig{fig:system_namespaces_bdd}.

\begin{figure}
	\centering
	\includegraphics[scale=0.65]{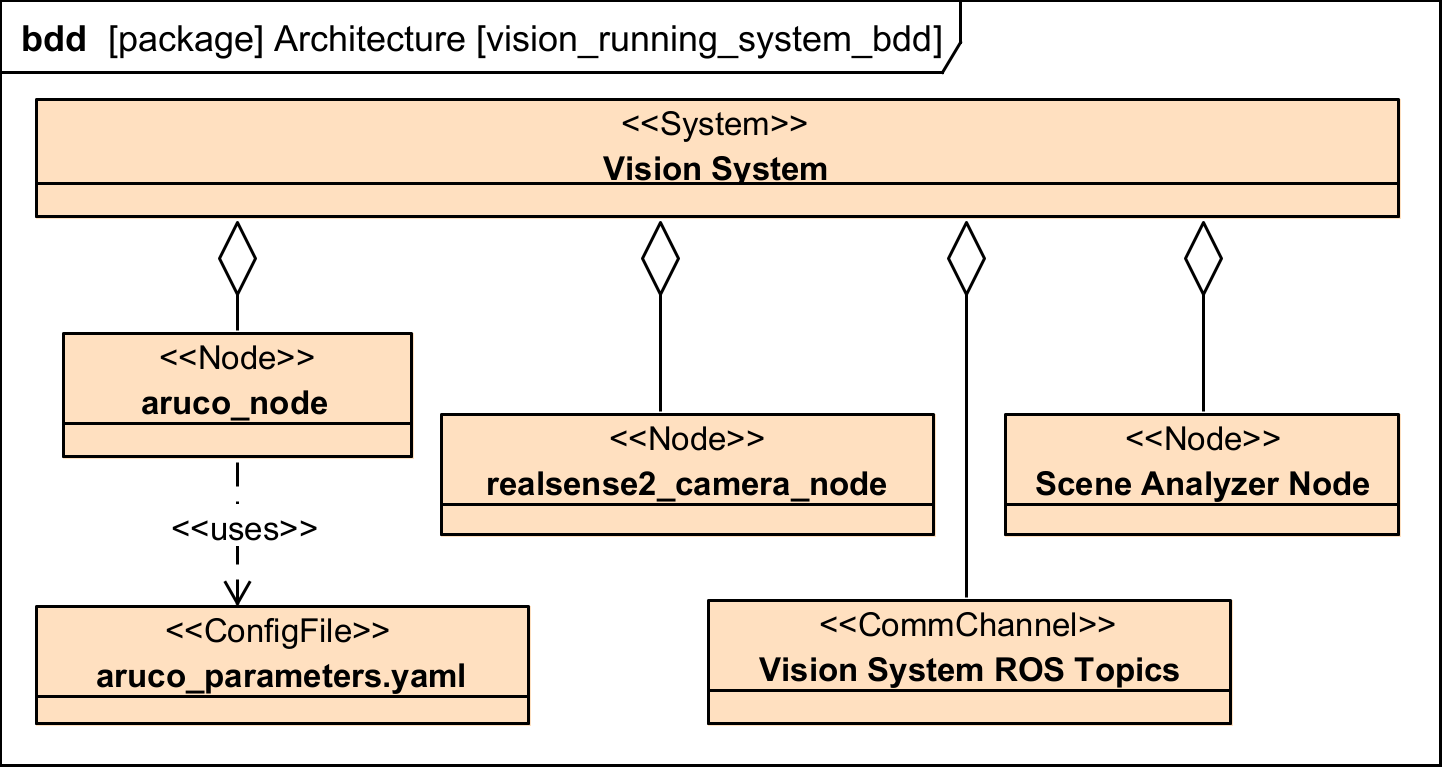}
	\caption{Structure of the \model{Vision System}.}
	\label{fig:vision_running_system_bdd}
\end{figure}

The \model{Vision System} components presented in \Fig{fig:vision_running_system_bdd} consist of three \solution{ROS~2} \stNode[][s] that exchange data by the \stTopic{} mechanism. The \rosNode{aruco\_node} \stNode{} is responsible for AR-marker detection. \stParameter[][s] from the \rosExePathName{aruco\_parameters.yaml} file are passed to this node. The \rosNode{realsense2\_camera\_node} communicates with the camera and publishes its acquired image stream. The \model{Vision System} also includes a~dedicated \model{Scene Analyser Node} to analyse observations from the environment. Based on the camera image, it determines whether the \model{Mobile robot} is ready to receive or return the cube.

\begin{figure}
	\centering
	\includegraphics[scale=0.59]{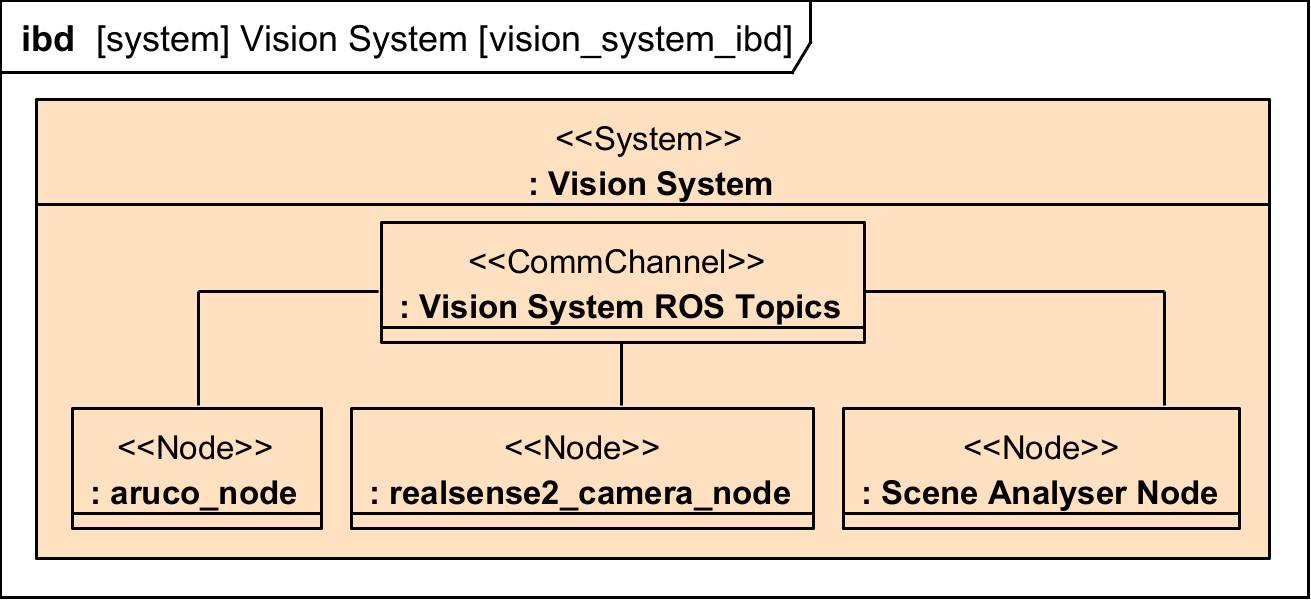}
	\caption{\diagram{Internal block diagram} of \model{Vision System}.}
	\label{fig:vision_system_ibd}
\end{figure}

The \diagram{internal block diagram} in \Fig{fig:vision_system_ibd} specifies that the \stNode{} components of the \model{Vision System} communicate with each other through the \model{Vision System ROS Topics}\stCommChannel{}.

\begin{figure}
	\centering
	\includegraphics[scale=0.65]{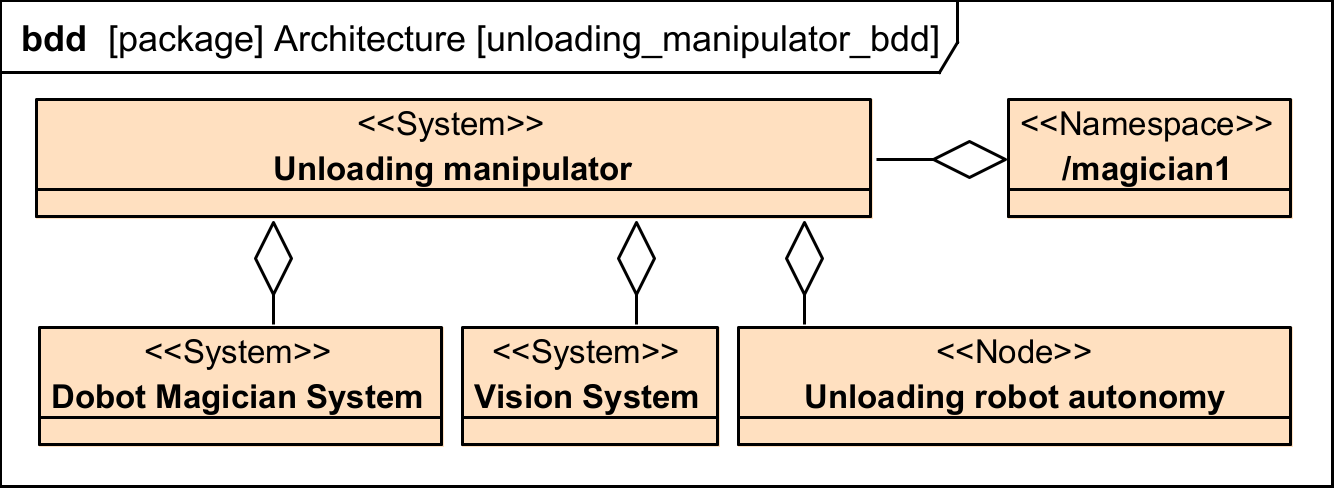}
	\caption{Composition of \stSystem[][s] in \model{/magician1} \stNamespace{}.}
	\label{fig:unloading_manipulator_bdd}
\end{figure}

The last one described will be the \model{Unloading manipulator} \stSystem{}, which is responsible for picking up cubes from the \model{Mobile robot}. Its structure is depicted in \Fig{fig:unloading_manipulator_bdd}. It is launched in the \rosNamespace{/magician1} \stNamespace{}, which prevents name conflicts with other \stCommChannel{} and \stRosCommCompon{} of this \stSystem{}. It consists of the \model{Dobot Magician System} manipulator controller, the \model{Vision System}, and the custom \model{Unloading robot autonomy} \stNode{} responsible for selecting the robot's behaviour in~response to external stimuli. Control systems of the other manipulators include the \model{Loading robot autonomy} \stNode{} and the \model{Supporting robot autonomy} \stNode{}, respectively.

\begin{figure}
	\centering
	\includegraphics[scale=0.65]{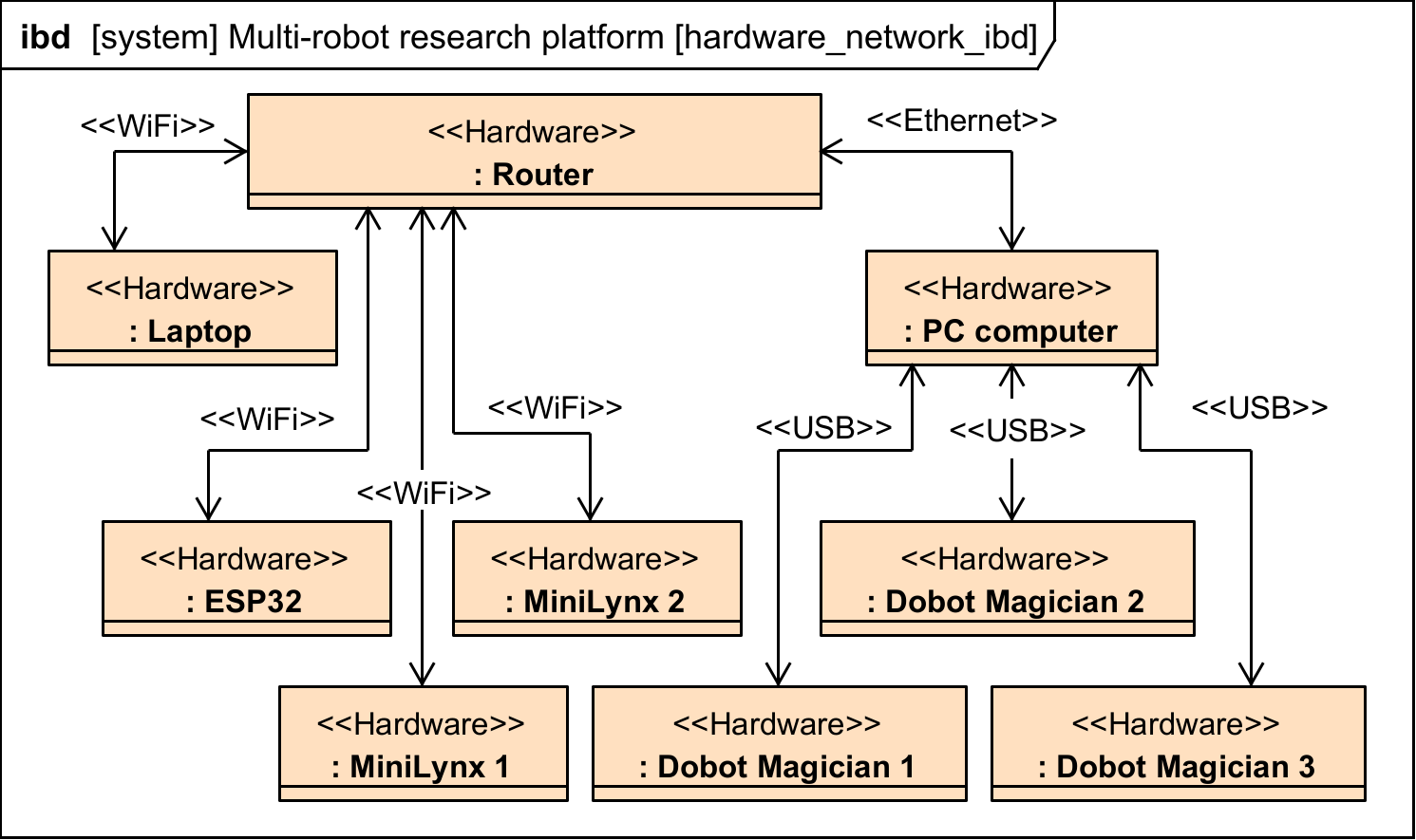}
	\caption{\stHardware{}-Network architecture of the multi-robot \stSystem{}.}
	\label{fig:hardware_network_ibd}
\end{figure}

As part of the \tagref{ACTV-DD} stage, it is also essential to plan how the \stHardware{}-Network architecture of the designed \stSystem{} will look (\Fig{fig:hardware_network_ibd}).

\subsection{Software/Hardware realisation \tagref{ACTV-SHR}}

The realisation of the \stSystem{} can be divided into \model{Software} and \stHardware{} parts. As for the \model{Software}, it was based on \solution{ROS~2} (Humble Hawksbill distribution \cite{Ros2Humble2022}). The high-level part of the software was written in Python. C~language was used to program the microcontroller.

In this work, we followed ROS Enhancement Proposals (REPs) \cite{ROS_REP}, primarily REP~144 and REP~135, to ensure compliance with established standards in \solution{ROS~2} development. REP~144 was followed to ensure a standardised package structure and consistent naming practices. Additionally, REP~135 was applied to standardise the use of \stNamespace{} and naming patterns, ensuring organised and unambiguous identification of \stNode{} and \stTopic{} in the \solution{ROS~2} ecosystem.

The \stHardware{} part required designing and printing 3D components (accessories for the \model{Manipulator} and \model{Mobile robots}, moving \model{Obstacles} and parts of the environment). Cameras had to be mounted on the \model{Manipulators}, as well as extenders for two-finger grippers that made it easier to grasp the cube. Two DC motors, a two-channel motor controller and an \model{ESP32} microcontroller were used to control the moving \model{Obstacles}.

\subsection{SubSystem realisation verification \tagref{ACTV-SSRVE}}

The SubSystem realisation verification \tagref{ACTV-SSRVE} stage involves the comparison of the compatibility of the realised architecture with its design. In the case of \solution{ROS}-based \stSystem{}, this is a check of the compatibility of names for each \stTopic{}, \stNode{}, \stNamespace{} and other \stRunSystemCompon[][s]. Because this is the SubSystem verification stage, the process should focus only on connections between \stNode[][s] within each \stSystem[Sub], but not on the entire \stSystem{} integration.

\solution{ROS~2} provides a wide range of tools, both graphical user interface (GUI) and command-line interface (CLI), enabling introspection of the running \stSystem{}. From GUI tools, it is worth using \rosNode{rqt\_graph} to visualise the relevant \stNode{} and \stTopic{} connections scheme. From the CLI tools (\stCLTool{}), it is convenient to choose \texttt{`ros2 node info <node\_name>`} and \texttt{`ros2 topic info <topic\_name>`} and confront it with the \stSystem{} design model.

At this stage, you should also verify the correctness of \stPackage{} names and whether each artefact and code element is placed in the prescribed \stRepository{}, \stPackage{} and \stWorkspace{}.

In the \stSystem{} described in this article, the abovementioned tools were used, which made it possible to successfully carry out the \tagref{ACTV-SSRVE} stage.

\subsection{SubSystem Validation Plan \tagref{ACTV-SSVAP}}

This stage involves planning various test scenarios to validate the operations of \stSystem{}. The outcome of this stage is a set of \diagram{behavioural diagrams}, which will be used in the following \tagref{ACTV-SSVA} stage of the proposed V-model procedure. The test scenarios are derived from the initial top-level \stRequirement[][s] (See \Fig{fig:general_req}).

\begin{figure}
	\centering
	\includegraphics[scale=0.65]{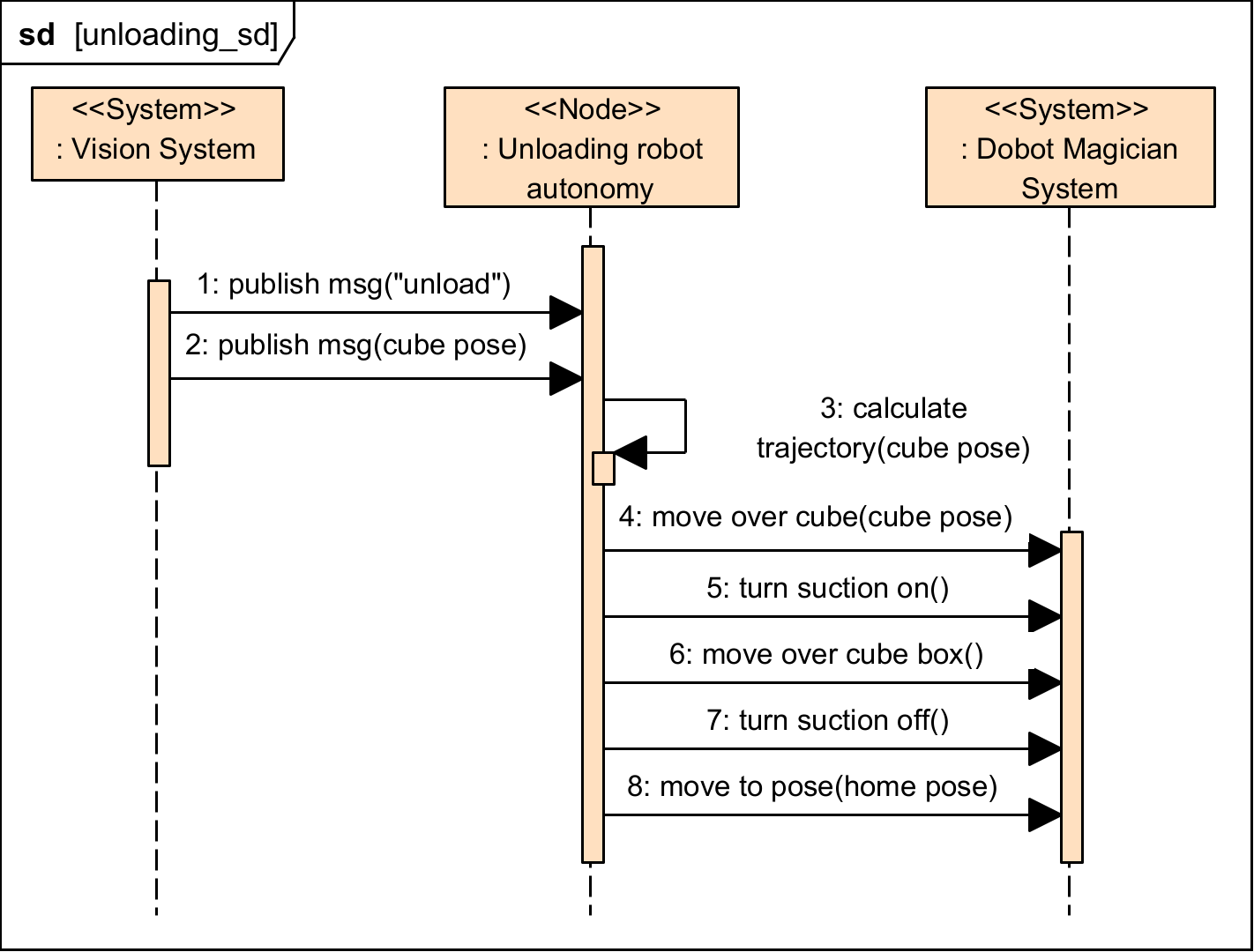}
	\caption{Sequence diagram for unloading a cube from a \model{Mobile robot} (video at 0:35): \href{https://vimeo.com/977486838\#t=35.112}{vimeo.com/977486838}.}
	\label{fig:unloading_sd}
\end{figure}

\Fig{fig:unloading_sd} shows the sequence of operations performed when picking up the cube from the robot. The \model{Unloading robot autonomy} \stNode{} will only initiate picking of the cube when the \model{Mobile Robot} stops moving. The picked cube is dropped into a~box standing on the left side of the \model{Unloading manipulator}. Finally, the \model{Unloading robot} \activity{moves to the base (home) pose} to observe the predefined area of the base with the camera.

\begin{figure}
	\centering
	\includegraphics[scale=0.65]{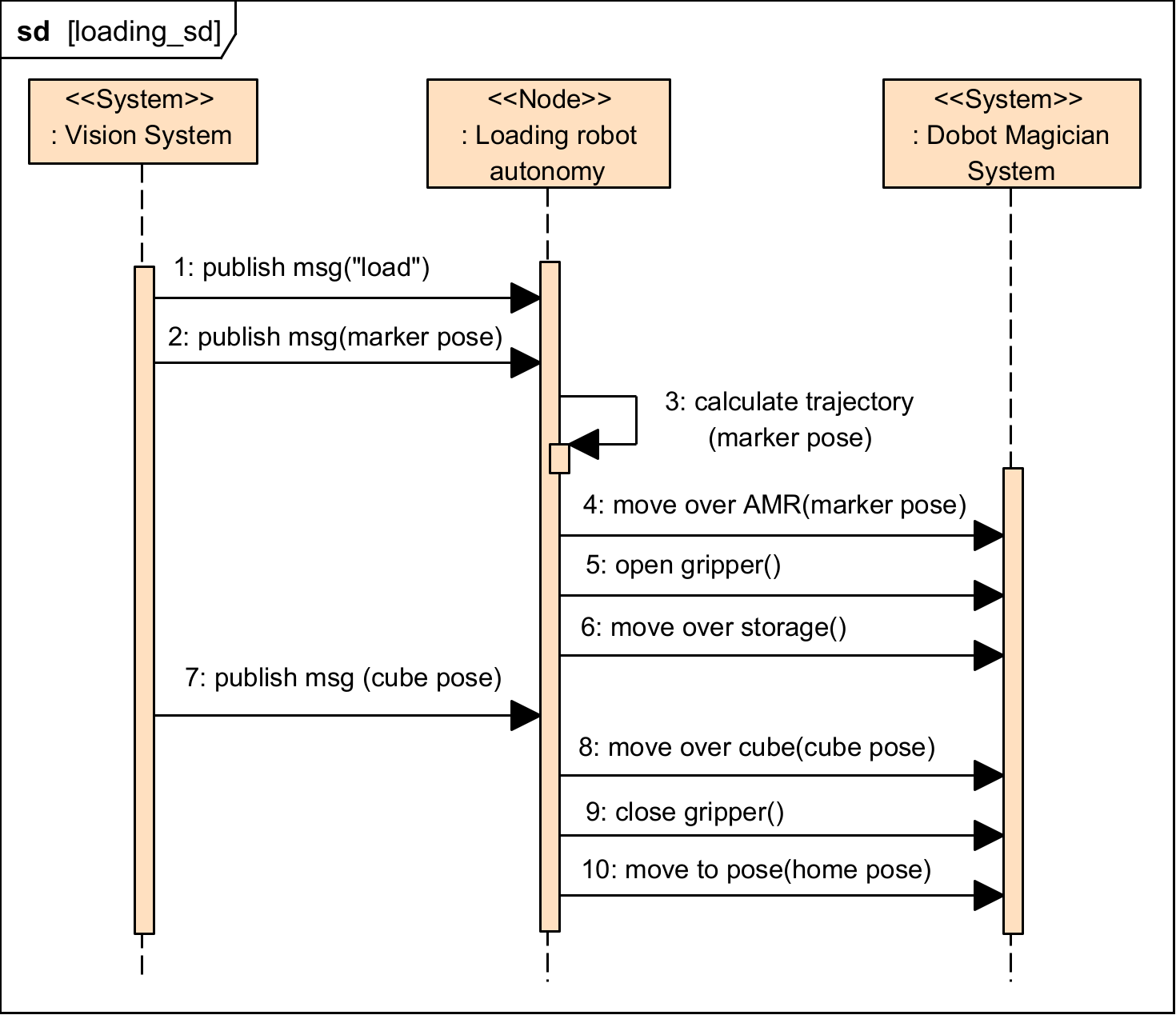}
	\caption{\diagram{Sequence diagram} for placing a cube on a \model{Mobile robot} (video at 0:29): \href{https://vimeo.com/977486838\#t=29.000}{vimeo.com/977486838} .}
	\label{fig:loading_sd}
\end{figure}

The sequence of operations for placing a cube on a \model{Mobile robot} is depicted in \Fig{fig:loading_sd}. The process follows these stages:
(1) The \model{Vision System} sends a load message to initiate the sequence;
(2) the \model{Loading robot autonomy} \stNode{} retrieves the \model{AR-marker} position attached to the \model{Mobile robot};
(3) it calculates an approach trajectory;
(4) the manipulator moves above the \model{Mobile robot};
(5) a signal is sent to open the gripper;
(6) the arm moves above the cube storage area;
(7) the \model{Vision System} provides the position of the target cube;
(8) the \model{Manipulator} moves above the cube and picks it up;
(9) the gripper closes;
(10) finally, the arm returns to a base observation pose, from which it monitors the board for the next incoming \model{Mobile robot}.

\begin{figure}
	\centering
	\includegraphics[scale=0.65]{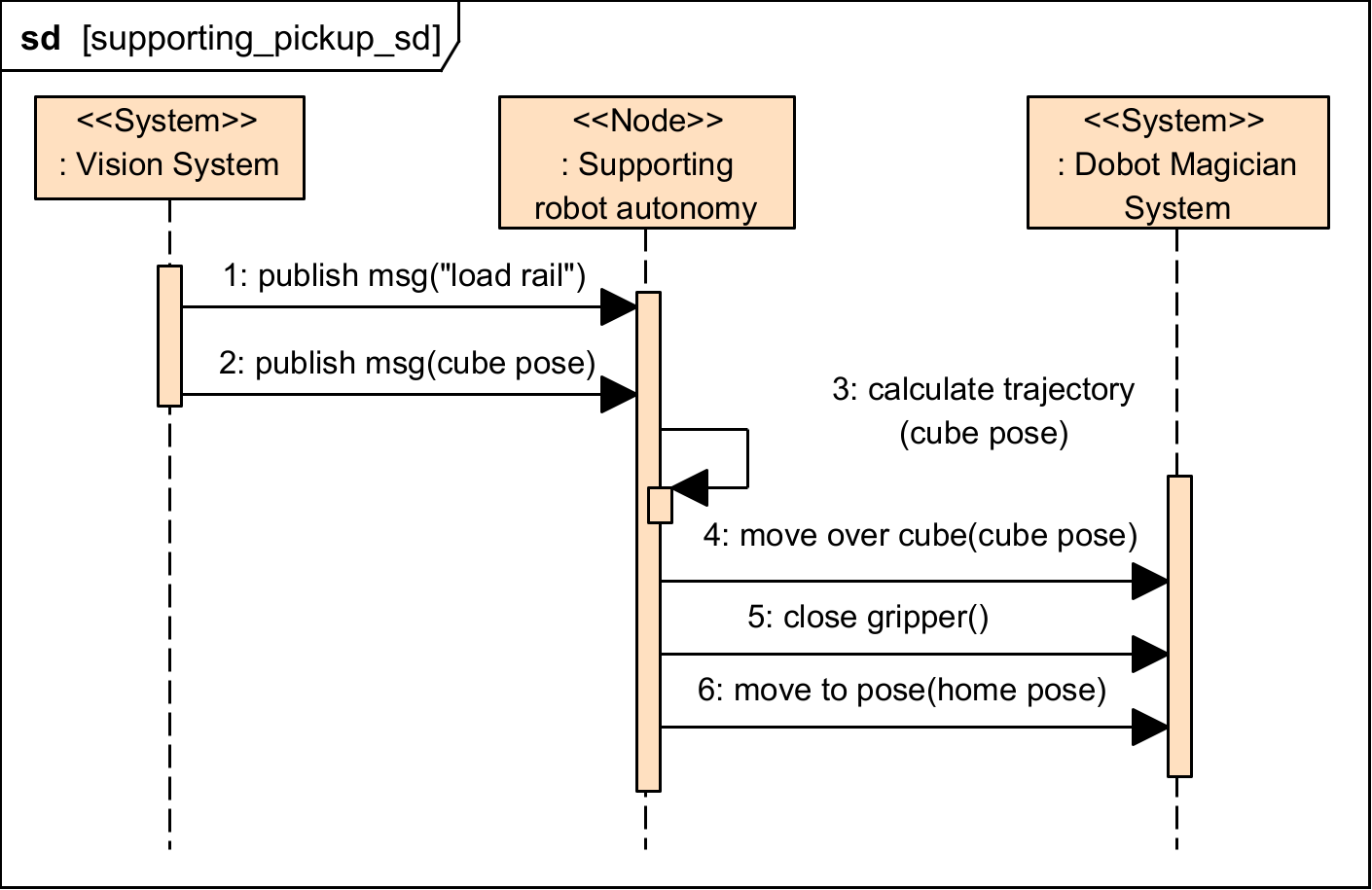}
	\caption{\diagram{Sequence diagram} for the \model{Supporting manipulator} picking up a cube (video at 2:08): \href{https://vimeo.com/977486838\#t=127.832}{vimeo.com/977486838} .}
	\label{fig:supporting_pickup_sd}
\end{figure}

\begin{figure}
	\centering
	\includegraphics[scale=0.65]{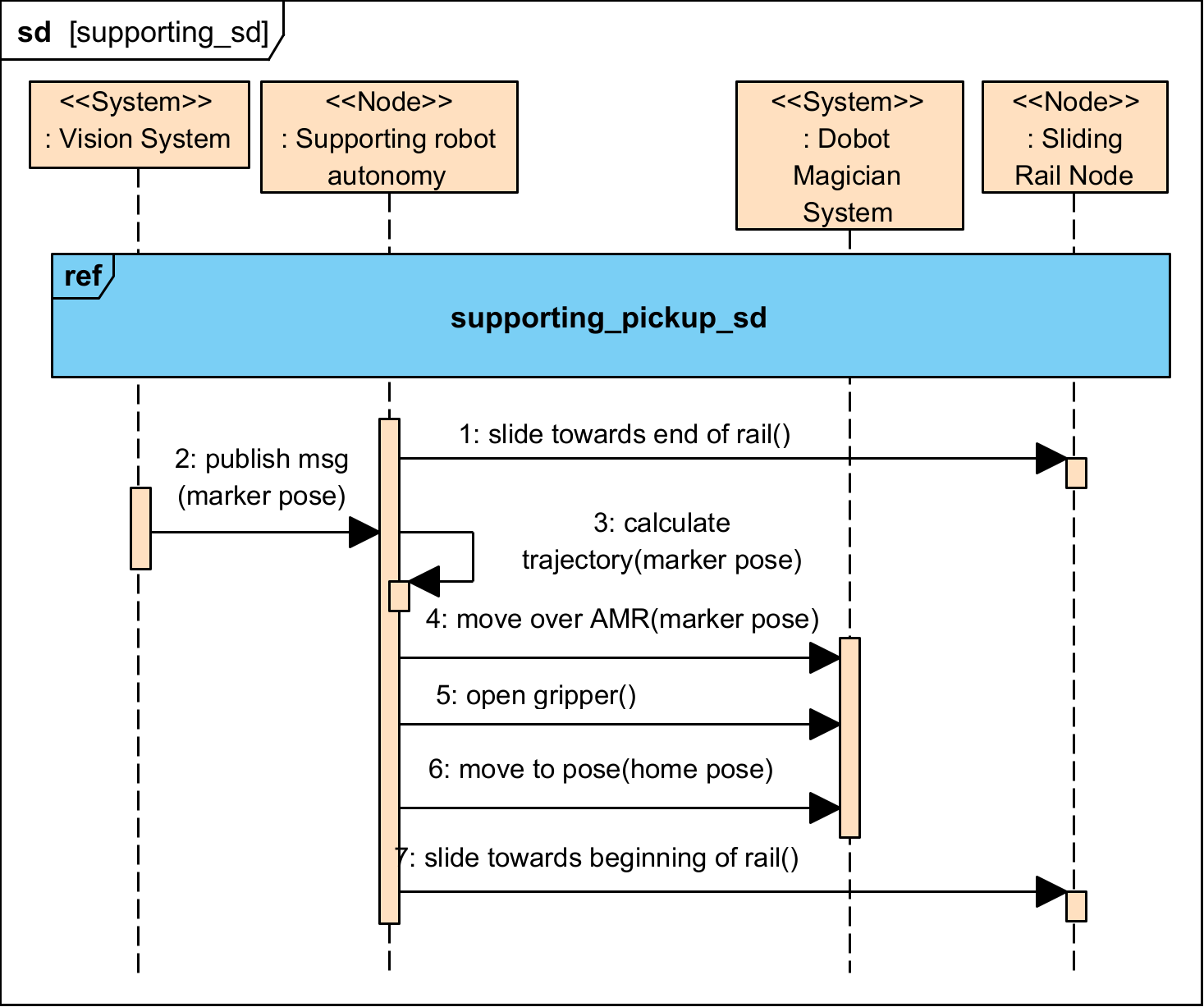}
	\caption{\diagram{Sequence diagram} for transporting a cube over an \model{Obstacle} (video at 2:08): \href{https://vimeo.com/977486838\#t=127.832}{vimeo.com/977486838} .}
	\label{fig:supporting_pickup_sd_2}
\end{figure}

The \diagram{sequence diagrams} shown in \Fig{fig:supporting_pickup_sd}~and~\Fig{fig:supporting_pickup_sd_2} depict the sequence of operations performed by the \model{Supporting manipulator}. The role of the \model{Supporting manipulator} is to transfer a~cube from one \model{Mobile robot} to another \model{Mobile robot} in a case environment is impassable. When the \model{Supporting robot autonomy} \stNode{} receives a~\rosMessage{"load rail"} \stMessage{} from the \model{Vision System}, the robot will pick up the cube from the \model{Mobile robot~1}. It will then start moving along the rail until it encounters a~\model{Mobile robot~2} on which it can put the cube down. \model{Sliding Rail Node} is part of the composition of the \model{Supporting manipulator} \stSystem{}.

\begin{figure}
	\centering
	\includegraphics[scale=0.65]{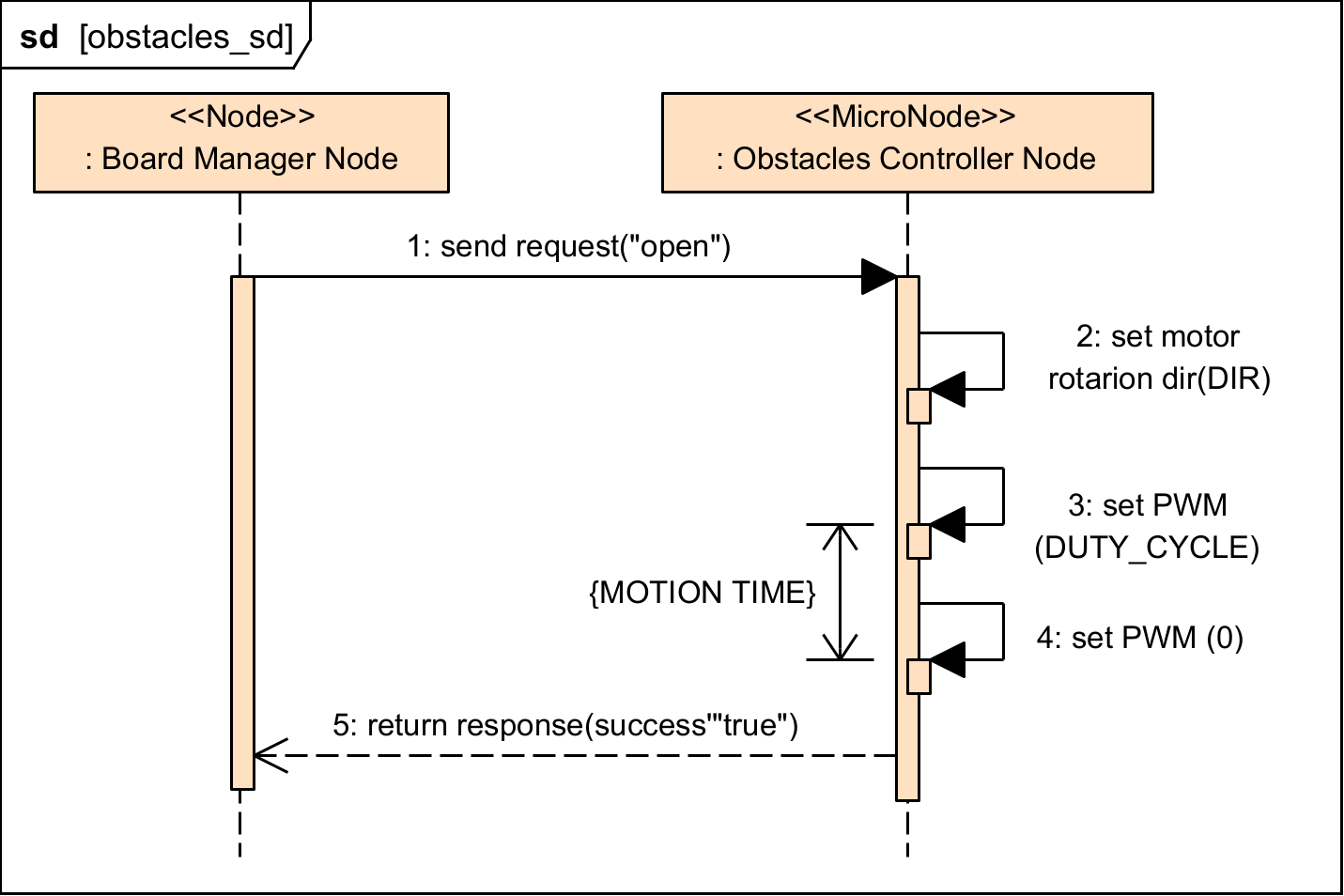}
	\caption{\diagram{Sequence diagram} of operations performed when moving an \model{Obstacle} (video at 1:24): \href{https://vimeo.com/977486838\#t=83.904}{vimeo.com/977486838} .}
	\label{fig:obstacles_sd}
\end{figure}

The sequence of operations performed by the \model{ESP32} microcontroller while moving the \model{Obstacle} is shown in diagram \Fig{fig:obstacles_sd}. Both the \model{Board Manager Node} \stNode{} and the \model{Obstacles Controller Node} \stMicroNode{} are composed into the \model{Obstacles} \stSystem{}.

\subsection{SubSystem realisation validation \tagref{ACTV-SSVA} }

Validation takes place at various levels of the V-model and confirms the functionality of individual components and the \stSystem{} as a whole. As part of the \stSystem[Sub] realisation validation \tagref{ACTV-SSVA}, its functionality is tested based on scenarios specified during the \tagref{ACTV-SSVAP} stage.

\begin{figure}
	\centering
	\includegraphics[width=\columnwidth]{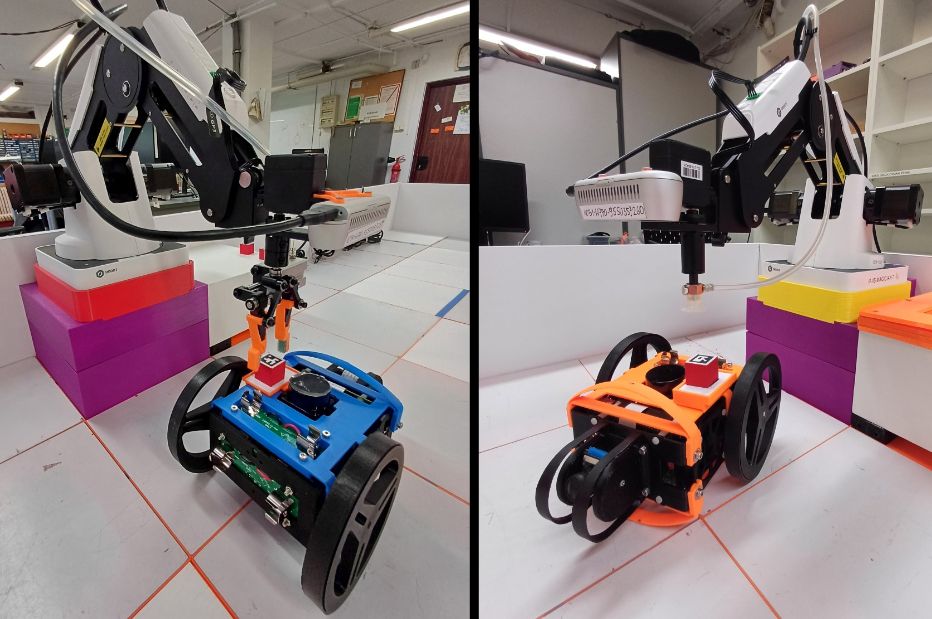}
	\caption{Cooperation of \model{Mobile robots} and \model{Manipulators}.}
	\label{fig:load_unload_img}
\end{figure}

\begin{figure}
	\centering
	\includegraphics[width=\columnwidth]{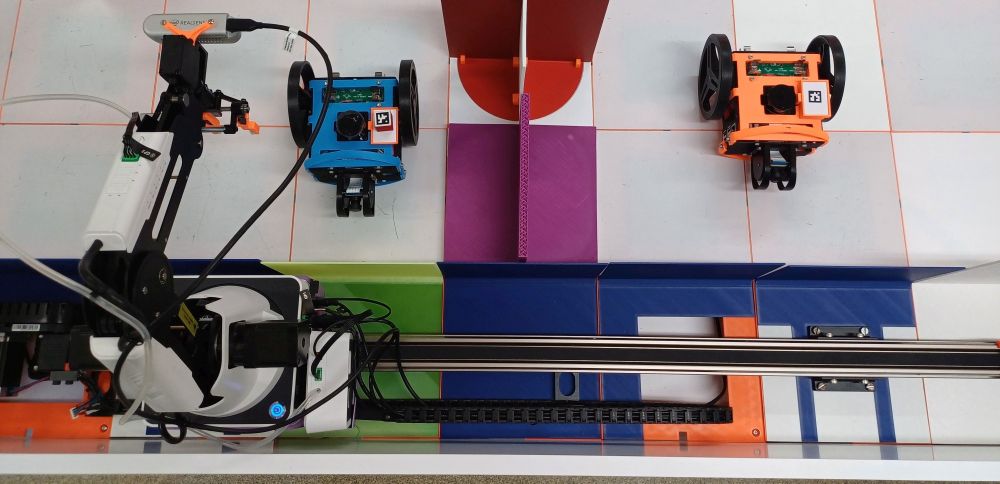}
	\caption{\model{Mobile robots} are waiting for support from the \model{Manipulator} mounted on the \model{Sliding rail}.}
	\label{fig:rail_support_cropped}
\end{figure}

\begin{figure}
	\centering
	\includegraphics[width=\columnwidth]{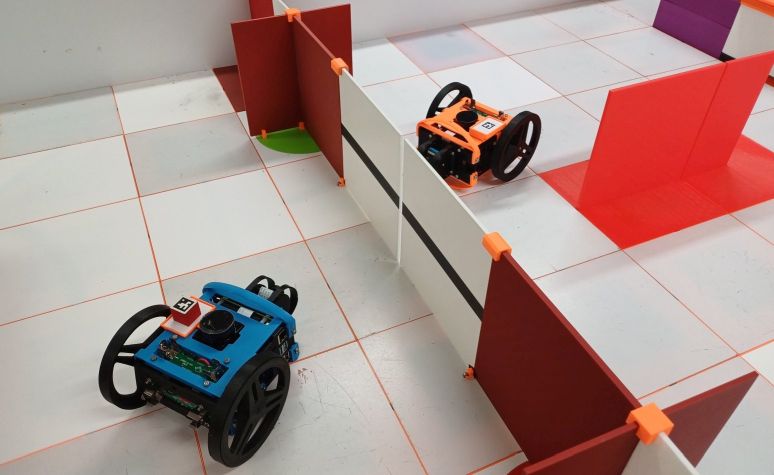}
	\caption{Closed gate preventing \model{Mobile robots} from driving to the other side of the \model{Board}.}
	\label{fig:closed_gate}
\end{figure}

The validation had been successful for all test scenarios. The most relevant fragments from the execution of test scenarios are presented below in the form of the following snapshots:
\begin{itemize}
    \item \model{Manipulator}s picking and placing the cube (\Fig{fig:load_unload_img});
    \item mounted on a sliding rail, the \model{Manipulator} supports \model{Mobile robots} (\Fig{fig:rail_support_cropped});
    \item \model{Mobile robots} are unable to continue the task because the \model{Obstacle} restricts passage (\Fig{fig:closed_gate});
\end{itemize}

\subsection{System Verification \tagref{ACTV-SRVE}}

At this stage, we verify integration between the \stSystem[Sub][s], focusing on the compatibility of the \stTopic{} and the interfaces that connect them. The same tools that were applied for \tagref{ACTV-SSRVE} are used at the higher \stSystem{} level.

\subsection{System Validation Plan \tagref{ACTV-SVAP}}
\begin{figure}
	\centering
	\includegraphics[scale=0.65]{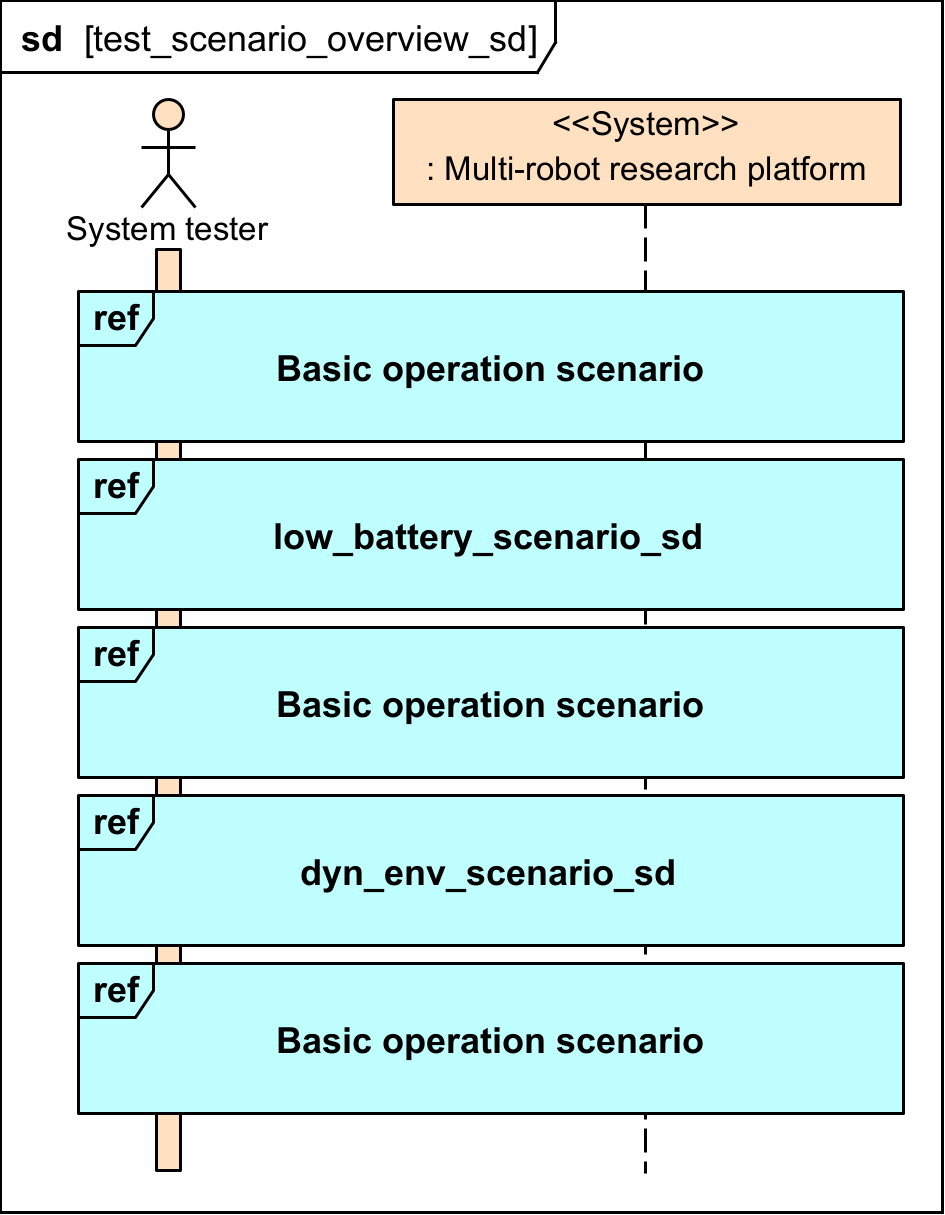}
	\caption{Full test scenario.}
	\label{fig:test_scenario_overview_sd}
\end{figure}

A~validation scenario to test the \stSystem{}'s performance is shown in \Fig{fig:test_scenario_overview_sd}. It consists of five parts; we will focus on the fourth part (the penultimate one) because it is the most elaborate and therefore the most representative.

\begin{figure}
	\centering
	\includegraphics[scale=0.65]{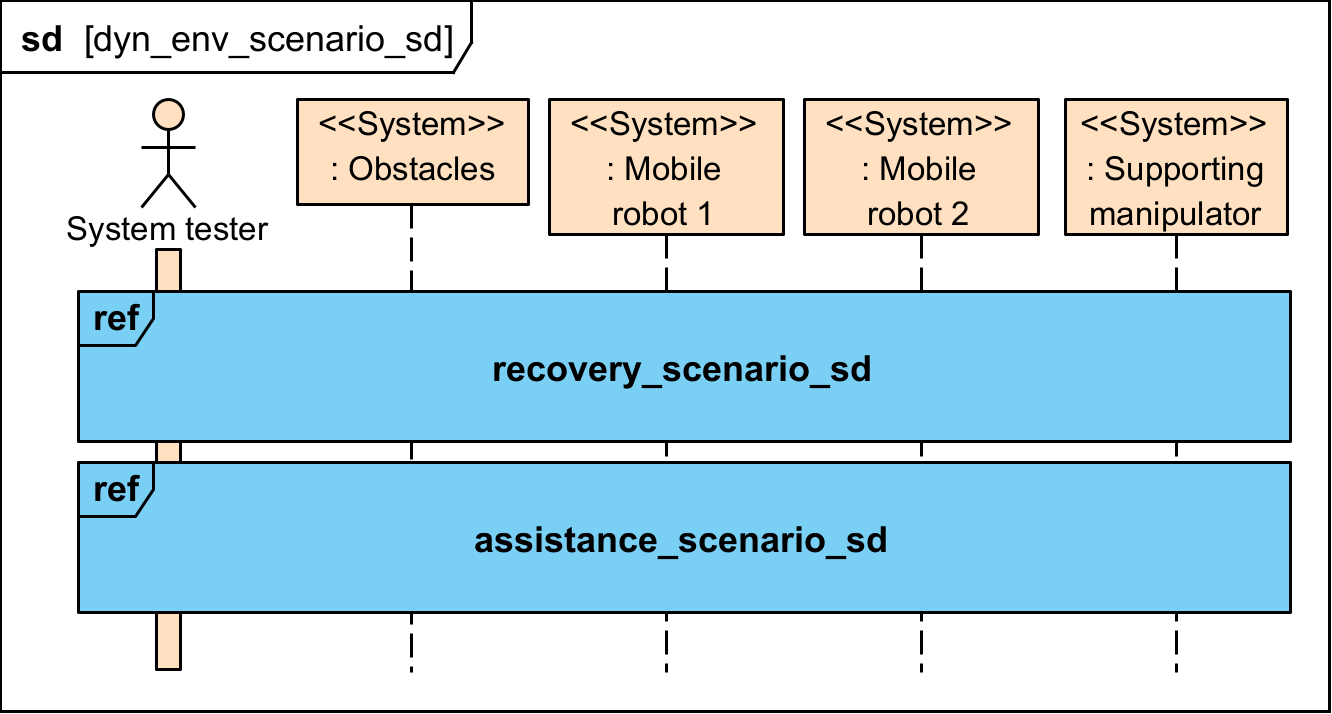}
	\caption{\diagram{Sequence diagram} considering the occurrence of changes in the environment.}
	\label{fig:dyn_env_scenario_sd}
\end{figure}

The scenario describing the operation of the \stSystem{} in a~changing environment consists of two sub-scenarios (\Fig{fig:dyn_env_scenario_sd}). The first describes the \model{Mobile robots}' reaction to the reconfiguration of the environment, and the second describes the role of the \model{Supporting manipulator}.

\begin{figure}
	\centering
	\includegraphics[scale=0.65]{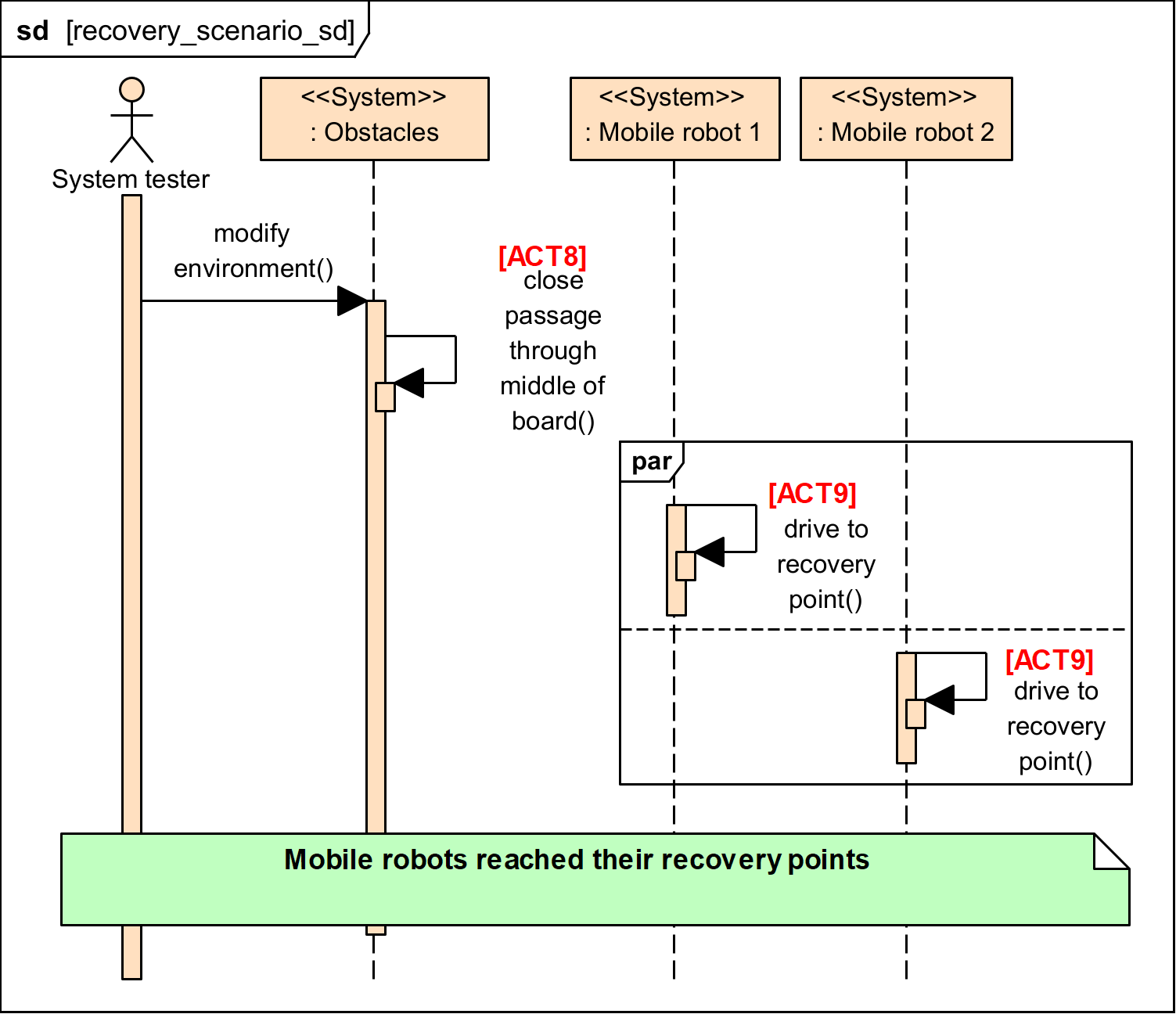}
	\caption{\diagram{Sequence diagram} describing the reaction of \model{Mobile robots} to changes in the environment (video at 1:24): \href{https://vimeo.com/977486838\#t=84.69}{vimeo.com/977486838} .}
	\label{fig:recovery_scenario_act}
\end{figure}

\begin{figure}
	\centering
	\includegraphics[scale=0.65]{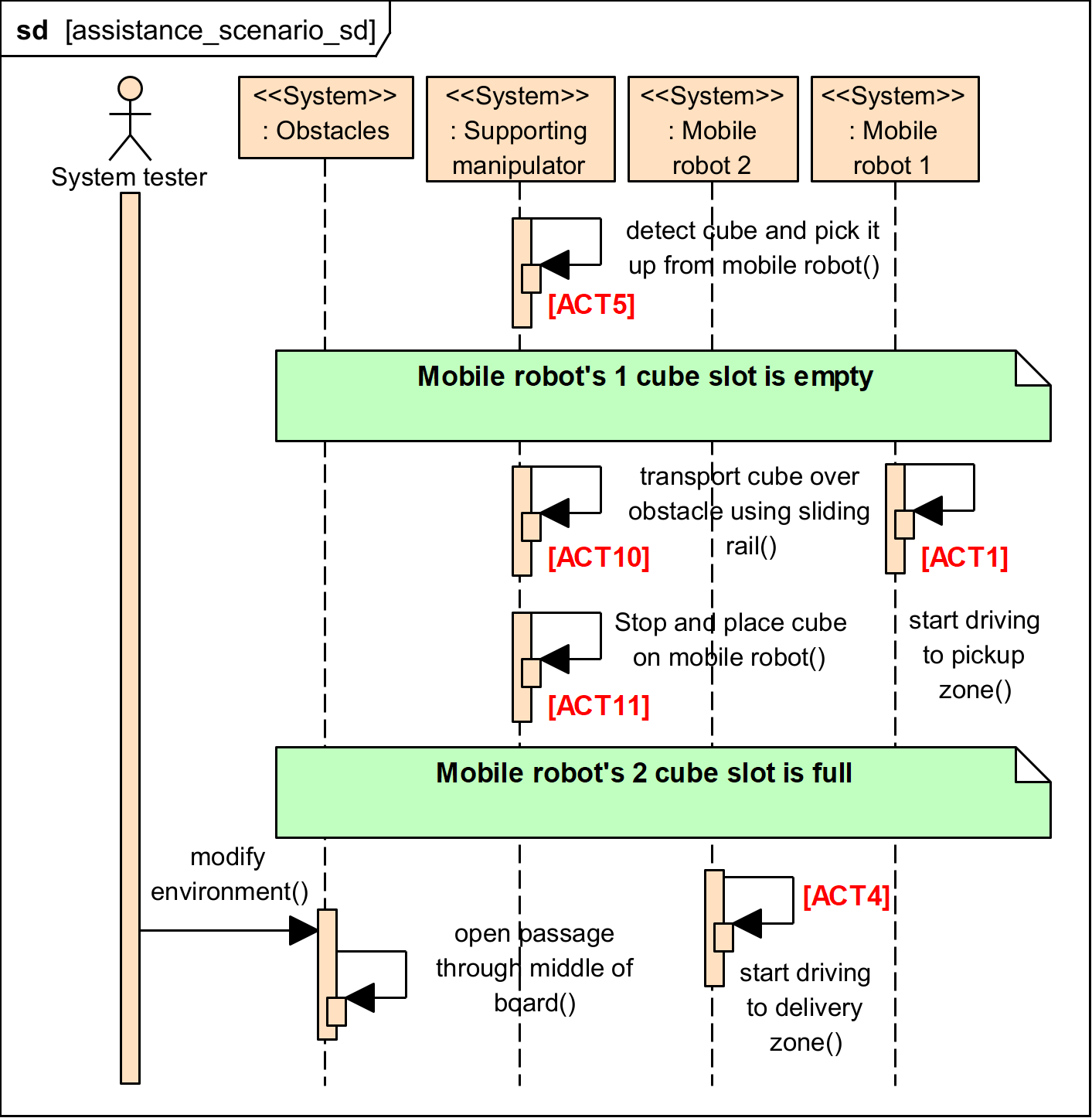}
	\caption{\diagram{Sequence diagram} representing the role of the \model{Supporting manipulator} in the cube transport task (video at 2:09): \href{https://vimeo.com/977486838\#t=129.624}{vimeo.com/977486838} .}
	\label{fig:assistance_scenario_act}
\end{figure}

\diagram{Sequence diagrams} describing the aforementioned sub-\-sce\-narios are shown in \Fig{fig:recovery_scenario_act} and \Fig{fig:assistance_scenario_act}. An element of validation was to annotate in the following diagrams the activities that were specified at the stage of conceptualising the \stSystem{} operation.

\subsection{System validation \tagref{ACTV-SVA}}

\stSystem{} validation \tagref{ACTV-SVA} is the final step of the proposed procedure. It involves checking the functionality of the \stSystem{} as a whole. Validation tests were carried out according to the test scenarios specified at the \tagref{ACTV-SVAP} stage, designed to validate the fulfilment of the \stRequirement{} collection described within the \tagref{ACTV-COFRD} stage.

The video recorded during the execution of the validation scenario is available at the link \url{https://vimeo.com/977486838}.

\section{Related Work}
\label{sec:related-work}

  \phantomsection
  \label{sec:focus-of-reasons}
  The realisation of robotic systems, including the \solution{ROS}-fo\-cu\-sed ones, has been shaped by the same technical challenges relevant today: hardware integration, modularity of components, real-time operation and scaffolding for the robotic system's lifecycle support.

  Due to the number of competence fields in a typical project, we cannot ignore the organisational problems that come to the foreground: steep learning curves for methods and tools, and, most importantly, the impact on team collaboration. Narrow, field-specific solutions are numerous, but they mainly cover domain-specific artefacts that contain too many details to reuse for system-level design. Excessive specificity must be abstracted in the general representation of the system for team-wise clarity. Still, the specific design artefacts are valuable and must be attached to the appropriate high-level representation points in the system's model.

  Also, we pay separate attention to the safety concerns among the examples and cases of certification compliance. With robotics increasingly deployed in regulated and safety-critical domains, system engineering now faces heightened requirements for traceable documentation and structured verification (e.g., ISO~13482, IEC~61508, UL~3300 \cite{ISO-13482,IEC-61508,UL-3300}). Despite the substantial challenges, this shift has pushed forward the development of MBSE-aligned frameworks, and some examples of systematic practice are arising.

  \phantomsection
  \label{sec:motivation-of-the-subjects}
  Given all the abovementioned traits diversity, no single approach can address them all comprehensively. Finding a good fit for each aspect of these diverse traits requires radically different methods by scale and type. We also consider marginal solutions, as they can mark the response to the problem or even the rise of a new trend. Adoption curves and the long-term viability of the methods are definitive for their uptake in the developer community, yet these aspects are rarely discussed in academic literature. This is why \textit{post-mortem} reflections, such as M.~Henning's highlight of the reasons for \solution{CORBA}'s decline \cite{Henning2008FallCORBA}, are exceptionally valuable.

  Our comparison, therefore, spans a wide range of entities, including frameworks, toolchains, metamodels, and methodologies, differing in type and granularity. Throughout this work, we collectively refer to them as \solution{engineering solutions} \taginline{ES} (or just \solution{solutions}, for brevity), following the established practice \cite{Casalaro2022ModelDriven,DeSilva2021SurveyMDE}.

  \phantomsection
  \label{sec:time-frame-motivation}
  We leave the earliest ad-hoc practices preceding the industry-wide methodologies out of the time scope. Such approaches rarely considered integration or lifecycle issues and retained a limited long-term impact \cite{Brugali2015MdseInRobotics}. Our review starts from the middleware era (e.g., \solution{CORBA} and related systems), when the focus turned to reproducible, scalable engineering.

  \phantomsection
  \label{sec:comparison-axis}
  To clarify our rationale for the \solution{MeROS} and V-model approach synthesis, we analyse these lineages and the engineering trade-offs that have shaped their development. This context grounds our methodological choices in the requirements that have emerged as robotics systems move from laboratory prototypes toward real-world, safety-critical deployment.

  \phantomsection
  \label{sec:structure-of-the-compared-representatives}
  Based on the findings of previous mapping studies \cite{Casalaro2022ModelDriven,DeSilva2021SurveyMDE}, we group the principal \solution{engineering solutions} into three categories:

  \begin{itemize}
      \item \textbf{Component-based Middleware:} solutions focused on distributed execution, modularity, and interface harmonisation, exemplified by \solution{Robot Operating System (ROS)}, \solution{OpenRTM}, \solution{OPRoS}, \solution{Orocos}, and \solution{SmartSoft}. \solution{ROS}, in particular, has become the baseline infrastructure for much of modern robotics, but lacks native support for system-level modelling, verification, or lifecycle management.
      \item  \textbf{Metamodel-driven Integration:} Transitional frameworks introduce explicit component metamodels and tool support without full system-level coverage (e.g., \solution{BCM}), which sets a bridge between code-centric and model-centric engineering \cite{Casalaro2022ModelDriven}.
      \item \textbf{MBSE-oriented Frameworks:} Full MBSE environments for system-level design, verification, and runtime adaptation (e.g., \solution{ISE\&PPOOA}, \solution{MROS}, \solution{MeROS}), linking abstract models to deployed systems \cite{Winiarski2023MeROS,Brugali2015MdseInRobotics}.
  \end{itemize}
  A summary of these solution categories and their primary representatives is provided in \Tab{tab:framework-comparison}.

\subsection{Component-based Middleware}
\label{subsec:related-works-middleware}
  The robotics development era before the Component-based Middleware surge had been a domain of unique solutions for the same tasks over and over again. Projects were often created from scratch, with all the functionality tightly coupled and served by monolithic solutions with minor reuse concerns. The component-based design surge, with its availability for a broad audience and shift of the world-view towards decoupling problems, isolating concerns and making them small enough to be solved by readily available tools, is the root of the benefits we admire in the modern robotics landscape.

\paragraph{CORBA, RTC}
\label{par:ROOTS_CORBA}
  In the late 1990s, the Object Management Group (OMG) introduced their Common Object Request Broker Architecture (\solution{CORBA}), a language- and platform-agnostic middleware for distributed objects. Due to its iconic "location transparency" feature, users could call the remote object (e.g., over the network) the same way as the local ones \cite{WangSchmidt2001}. OMG-led standardisation efforts directly shaped lots of early solutions in robotics software. As a culmination of these efforts, it standardised the Robotic Technology Component (RTC) specification, based on top of \solution{CORBA} Component Model (CCM) and formalised in IDL, with UML often used for modelling design-level structure, it defined reusable, platform-independent robotics components with explicit execution semantics \cite{Kotoku2006RTC}. However, as Henning observes, committee-driven architectures, as in the case of CORBA, showed us that they are prone to ecosystem stagnation \cite{Henning2008FallCORBA}. Despite good initial results, with time, they fall back due to slow adaptation; without continual developer feedback, they risk becoming too rigid or complex for daily use, leaving no slack for adopting fresh trends.

  \solution{CORBA}-based solutions deserve credit for the effective spread of design modularity, which we are taking for granted, as well as for concepts of role-based components, life-cycle states, and controlled execution flows. Although the global influence of \solution{CORBA} came to its dusk, it is still present in specific robotics contexts: in domains of real-time systems and long-lasting installations. Such solutions as \solution{OpenRTM} and \solution{OPRoS} extend this lineage, whereas frameworks like \solution{Orocos} and \solution{BCM} have selectively inherited only specific concepts.

\paragraph{OpenRTM}
\label{par:OpenRTM}
  \solution{OpenRTM-aist}, developed by the National Institute of Advanced Industrial Science and Technology (AIST, Japan), is an open-source solution that implements the OMG RT-Middleware specification, building directly on the \solution{CORBA} Component Model and the RTC standard \cite{Kotoku2006RTC,AndoSuehiro2008}. Software components, termed RT-Components, are defined by standardised data and service ports, explicit execution contexts, and formal life-cycle states. This structure supports modular reuse, run-time introspection, and cross-language deployment. AIST also provides development tools, such as RTCBuilder and RTSystemEditor, to automate code generation and enable graphical system configuration. In practice, \solution{OpenRTM} is among the most complete open implementations of the \solution{CORBA} component approach for robotics.

  Variants targeting safety-critical domains, such as \solution{RTMSafety}, integrate SysML modelling with IEC~61508/61499 function blocks, providing partial traceability and basic support for functional safety \cite{HanaiRtComponent2012}. The same middleware is used by \solution{OpenHRP3}, which combines real-time robot control with physics-based simulation \cite{Kanehiro2004OpenHRP3}. The strengths and current limitations of \solution{OpenRTM}, especially with regard to engineering life-cycle support, are summarised in the \Tab{tab:framework-comparison}.

\paragraph{OPRoS}
\label{par:OPRoS}
  The \solution{Open Platform for Robotic Services (OPRoS)}, developed by the Electronics and Telecommunications Research Institute (ETRI, Korea), is an open-source framework extending \solution{CORBA} and the RTC specification. It provides an integrated, IDE-based toolchain for component-oriented robotics development \cite{JangLee2010OPRoS,Han2012OPRoS}. The graphical authoring and composition tools generate XML deployment profiles for a dedicated execution engine. An embedded simulator enables verification of system behaviour in the loop. A safety-oriented variant connects RTC components to IEC~61508/61499 function blocks, adding basic life-cycle traceability. \solution{OPRoS} demonstrates how RTC-based middleware concepts can be extended into a model-driven development environment.

\paragraph{CLARAty, SmartSoft}
\label{par:CLARAty}
  The Coupled-Layer Architecture for Ro\-bo\-tic Autonomy (\solution{CLARAty}), developed at NASA Jet Propulsion Laboratory (JPL, USA), was designed to support modularity and software reuse in planetary exploration missions \cite{NesnasVolpe2001}. Its two-layer design separates a Functional layer for hardware abstraction from a Decision layer for high-level autonomy, decoupling device control from deliberative reasoning and enabling deployment on heterogeneous platforms. \solution{SmartSoft}, developed at the Research Institute for Applied Knowledge Processing (FAW, Germany), focuses on predictable component behaviour via declarative coordination patterns and service-oriented communication \cite{SchlegelWorz1999}.

  Although both frameworks advanced middleware formalisation, neither integrates model-based systems engineering artefacts, such as requirement traceability or V-model alignment. Their adoption has remained limited, particularly for projects requiring life-cycle modelling or SysML-based tool support. For these reasons, they are omitted from detailed comparison, but they are recognised as important conceptual forerunners in the evolution of robotics software frameworks.

\subsubsection{Real-Time Control Frameworks}
\label{subsubsec:related-works-real-time}

The desire to bring robotics solution into areas of critical applications, with advancements in embedded computing resources, has helped tackle the domain of real-time interactions. Middleware in this category focuses on predictable execution and minimal latency, accepting reduced support for architectural abstraction or life-cycle integration as a trade-off.

Examples include \solution{TAO} \cite{Schmidt1998TAO}, a real-time \solution{CORBA} implementation developed at Washington University in St. Louis (USA), and \solution{ICE}(Internet Communications Engine), produced by the company ZeroC (USA) as a lightweight alternative. Both were released as open-source software, encouraging their adoption in academic robotics. Their primary focus on communication performance and interface abstraction limited their applicability to complex life-cycle management or model-based engineering.

In parallel, such frameworks designed for embedded control: \solution{Open Robot Control Software (Orocos)}, developed at KU Leuven (Belgium) \cite{Bruyninckx2001Orocos}, \solution{GenoM3}, developed at the Laboratory for Analysis and Architecture of Systems of the French National Centre for Scientific Research (LAAS-CNRS, France)~\cite{Mallet2010GenoM3}, and the \solution{BRICS Component Model (BCM)}, developed as part of the BRICS project (Germany)~\cite{BruyninckxKlotzbucher2013BCM}---prioritised modularity, run-time isolation, and execution fidelity. These solutions supported reliable deployment and operational safety, but only offered basic mechanisms for traceability, life-cycle modelling, or broad toolchain integration.

An intermediate example is \solution{Orca}, developed at the University of New Brunswick (Canada) \cite{BrooksKaupp2005}, which was built on \solution{ICE} and introduced composable interfaces and deployment semantics. Although no longer maintained, \solution{Orca} influenced subsequent frameworks with its approach to modularity and lightweight integration. Formal life-cycle modelling and verification, however, were not within its principal scope.

\paragraph{Orocos}
\label{par:Orocos}
The \solution{Open Robot Control Software (Orocos)} project \cite{Bruyninckx2001Orocos,Bruyninckx2003RealTime}, initiated at KU Leuven (Belgium), addresses deterministic scheduling and modular composition in real-time robotics. Its Real-Time Toolkit (\solution{RTT}) implements a component-based architecture that separates structure and function, event-driven execution, and low-latency control. \solution{Orocos} avoids enforcing a fixed system structure, allowing use across various applications, including servo control, simulation, and sensor integration. The design patterns developed in \solution{Orocos} informed the subsequent \solution{BRICS Component Model (BCM)}, which added formal modelling layers to support run-time modularity and deterministic operation.

\paragraph{GenoM / GenoM3}
\label{par:GenoM}
\solution{GenoM}~\cite{Mallet2002GenoM} was developed at LAAS-CNRS (France) to support modular execution in embedded autonomous robots. The original design organises sensing, control, and actuation into lightweight, service-based components. Its later development into \solution{GenoM3} introduced a formal specification layer and also supported code generation and component behaviour analysis. This structure allows system verification and design validation to be integrated with real-time execution, although it does not rely on standard SysML/UML tooling. \solution{GenoM3} thus provides a practical form of system modelling adapted to embedded robotics constraints.

Together, these frameworks represent a stage of robotics software engineering where formal architectural models and standardisation were introduced to improve reliability and modularity, but with limited support for comprehensive engineering life-cycle management or system-level validation.

\subsection{Metamodel-driven Integration}
\label{subsec:metamodel-driven}

\paragraph{BCM}
\label{par:BCM}
Within the \textit{Metamodel-driven Integration} category, the BRICS Component Model (\solution{BCM}) \cite{BruyninckxKlotzbucher2013BCM} represents joined efforts by Europe to formalise component-based robotics design, while remaining grounded in established runtime frameworks such as \solution{Orocos}. Developed under the BRICS initiative (Best Practice in Robotics), \solution{BCM} introduced a formal metamodel and a coordination schema covering the ``5C'' concerns: \textit{Computation}, \textit{Communication}, \textit{Coordination}, \textit{Configuration}, and \textit{Composition}. This structure supported platform-independent specification of component interfaces and behaviour, with transformation rules enabling \textit{integration into existing toolchains}, including \solution{Orocos} and \solution{ROS}. Unlike full MBSE environments, it lacked complete system-level lifecycle coverage and mature tooling. This transitional scope, which involves explicit metamodeling without full lifecycle breadth, is the defining criterion that Casalaro et al. \cite{Casalaro2022ModelDriven} use for the \textit{Metamodel-driven Integration} category.

\subsection{MBSE-oriented Frameworks}
\label{subsec:mbse-oriented}

The final lineage comprises Model-Based Systems Engineering (MBSE) frameworks, which integrate system-level modelling, requirements traceability, and lifecycle management into the robotics development process \cite{Brugali2015MdseInRobotics,Casalaro2022ModelDriven,MadniSievers2018}.

\paragraph{ISE\&PPOOA}
\label{par:ISEPPOOA}
\solution{ISE\&PPOOA} is a model-based systems engineering methodology that combines breadth-first requirements modelling with layered architecture design.
The method favours semi-formal diagrams and design heuristics to support modelling discipline in early system design. It extends SysML with \solution{PPOOA}'s port- and service-based constructs to model both software and system-level views of real-time systems. The method emphasises functional decomposition, hierarchical refinement, and modular design of software and hardware. While it does not support executable simulation or runtime testing, it enables traceability across stakeholder goals, system functions, and implementation artefacts. It has been applied in collaborative robotics and UAV domains \cite{Hernandez2017Collab}, with tool support focused on structure, reusability, and early-stage modelling discipline.

\paragraph{AutomationML, RobotML}
\label{par:AutomationML_RobotML}
\solution{AutomationML}, originally developed for industrial automation, has been extended to robotic domains via Semantic Web integration and code generation \cite{Hua2016AutomationML}. However, its lack of behavioural abstraction and SysML support limits its applicability in model-based system design, and it is not included in the lifecycle comparison.

\solution{RobotML} \cite{Dhouib2012RobotML} introduced a UML-profile-based DSL for robotic system modelling, simulation, and deployment targeting multiple middleware supported by a Papyrus-based toolchain and designed for deployment across platforms such as \solution{ROS} and \solution{Orocos}. Despite a rich domain model, it remains prototypical and has not been widely adopted. These frameworks highlight the variety of modelling approaches explored in robotics, but also underscore the difficulty of sustaining full lifecycle support and verification, gaps that more recent frameworks like \solution{MeROS} aim to address.

\paragraph{MROS (RobMoSys)}
\label{par:RobMoSys}
The RobMoSys was a European H2020 project promoting contract-based modelling and separation of concerns in robotics software. It introduced metamodels for roles, interfaces, and component composition, supported by tools such as SmartMDSD and Papyrus4Robotics. These enabled early-stage validation, component reuse, and structured integration workflows.
\solution{MROS}\cite{Silva2023MROS}, developed within this effort, extended the architecture to support runtime reconfiguration via ontological meta-control and variation point selection \cite{Bozhinoski2022Mros}. Demonstrations included adaptive planning and runtime monitoring \cite{Colledanchise2021Formalizing}. While \solution{MROS} provided an important lifecycle modelling tool, it remained peripheral to mainstream \solution{ROS} development. Still, its principles contributed to ongoing interest in model-centric robotics design.

\paragraph{PDD}
\label{par:PDD}
Property-Driven Design (\solution{PDD}) \cite{Brambilla2015PDD} focuses on swarm robotics and uses system-level properties to guide model refinement and verification. It avoids SysML/UML and uses formal methods (e.g. checking to validate emergent behaviour properties in swarm systems) with lightweight tooling. Although narrow in scope, \solution{PDD} demonstrates how behavioural verification can be integrated into iterative modelling workflows---a direction \solution{MeROS} generalises.

\paragraph{Positioning MeROS}
\label{par:MeROS}
The \solution{MeROS} ecosystem does not attempt to redefine model-based robotics but builds directly on the structure and lessons of earlier efforts. It follows a minimal, standards-aligned approach, grounded in SysML, to address a longstanding gap: linking architecture, behaviour, and runtime within a traceable, verifiable V-model process. While previous frameworks introduced valuable concepts, many remained fragmented or faded from practice. \solution{MeROS} stays close to the \solution{ROS} conventions while aiming to restore lifecycle consistency, which is often missing from open-source workflows. It is not a reinvention, but a practical synthesis of what has proven effective.

\vspace{0.5em}
\noindent
The detailed comparison of these model-based frameworks is provided by \Tab{tab:framework-comparison} in \Sec{sec:discussion}, supported by the discussion of their tradeoffs and implications.

\section{Discussion}
\label{sec:discussion}

\subsection{Motivation and Scope of Comparison}
\label{subsec:discussion-motivation}

In the previous \Sec{sec:related-work}, we traced how the development of robotics software systems has followed different lineages: from component-based middleware, mentioned a separate real-time-focused branch prioritising execution control, to the most mature MBSE-oriented frameworks advocating for lifecycle structure.
As far as the \solution{solutions} (see the \tagref{ES}) reviewed here and put in the context of this work were selected despite their scale, \Tab{tab:framework-comparison} situates them along a scale-agnostic axis of attention to the design stages: from conceptualising stakeholder requirements to subsystem-level verification in-depth. This mapping brings up how core architectural priorities often result in blind spots. Some solutions focus on the early-phase design formalisation, and others focus on runtime behaviour, but provide a thorough solution that can help justify the cost of extra procedures and tools; it is a long game with lots of endeavours and errors, and the best we can do is to collect the lessons of other efforts.
\Tab{tab:framework-comparison} synthesises these observations, mapping the lifecycle coverage of each framework and underscoring the methodological gaps that the \solution{MeROS} ecosystem seeks to resolve.

% MeROS_VModel_HeROS-RAS-Preprint-Submission.tex

{

\newcommand{\Ym}{\checkmark}
\newcommand{\Xm}{\xmark}
\newcommand{\Nm}{$\blacktriangle$}

\begin{table*}[t]
  \centering

  \caption{Comparison of approaches for robotic systems development.}
  \label{tab:framework-comparison}

  \scriptsize

  \resizebox{\textwidth}{!}
  {

  \begin{tabular}{
      >{\centering\arraybackslash}m{2.8cm}
      *{15}{>{\centering\arraybackslash}m{0.6cm}}
  }

  \toprule
  Source / Target
  & \rotatebox[origin=l]{90}{Code\&fix}
  & \rotatebox[origin=l]{90}{Concept plan}
  & \rotatebox[origin=l]{90}{System reqs}
  & \rotatebox[origin=l]{90}{System design}
  & \rotatebox[origin=l]{90}{System V\&V}
  & \rotatebox[origin=l]{90}{SubSys reqs}
  & \rotatebox[origin=l]{90}{SubSys design}
  & \rotatebox[origin=l]{90}{SubSys V\&V}
  & \rotatebox[origin=l]{90}{HW reqs}
  & \rotatebox[origin=l]{90}{HW design}
  & \rotatebox[origin=l]{90}{HW V\&V}
  & \rotatebox[origin=l]{90}{SW reqs}
  & \rotatebox[origin=l]{90}{SW design}
  & \rotatebox[origin=l]{90}{SW V\&V}
  & \rotatebox[origin=l]{90}{SysML/UML}
  \\

  \midrule

  \multicolumn{16}{l}{\textbf{Component-based:}}
   \\

  \textit{\textbf{Baseline:}}~\solution{ROS~/~ROS 2}
  & \cmark
  & \xmark
  & \xmark
  & $\blacktriangle$ / \xmark
  & $\blacktriangle$ / \xmark
  & \xmark
  & $\blacktriangle$
  & $\blacktriangle$ / \xmark
  & \xmark
  & \xmark
  & \xmark
  & $\blacktriangle$ / \xmark
  & $\blacktriangle$
  & $\blacktriangle$ / \xmark
  & \xmark
  \\
  \solution{OpenRTM}~+~\solution{RTMSafety} (Ando~\cite{AndoSuehiro2008},~Hanai~\cite{HanaiRtComponent2012})
  & \xmark
  & \xmark
  & $\blacktriangle$
  & \cmark
  & $\blacktriangle$
  & $\blacktriangle$
  & \cmark
  & \xmark
  & \xmark
  & \xmark
  & \xmark
  & $\blacktriangle$
  & \cmark
  & \xmark
  & \cmark
  \\
  \solution{OPRoS}~(Jang~\cite{JangLee2010OPRoS})
  & \cmark
  & \xmark
  & $\blacktriangle$
  & \cmark
  & $\blacktriangle$
  & $\blacktriangle$
  & \cmark
  & $\blacktriangle$
  & \xmark
  & \xmark
  & \xmark
  & $\blacktriangle$
  & \cmark
  & $\blacktriangle$
  & \xmark
  \\

  \solution{Orocos}~(Bruyninckx~\cite{Bruyninckx2003RealTime})
  & \cmark
  & \xmark
  & $\blacktriangle$
  & $\blacktriangle$
  & \xmark
  & \xmark
  & \cmark
  & \cmark
  & \cmark
  & \cmark
  & \xmark
  & \cmark
  & \cmark
  & \cmark
  & \xmark
  \\
  \solution{GenoM3}~(Mallet~\cite{Mallet2010GenoM3})
  & \xmark
  & \xmark
  & $\blacktriangle$
  & \cmark
  & \cmark
  & \cmark
  & \cmark
  & \cmark
  & \xmark
  & \xmark
  & \xmark
  & $\blacktriangle$
  & \cmark
  & \cmark
  & \xmark
  \\

  \arrayrulecolor{gray}\hline\arrayrulecolor{black}

  \multicolumn{16}{l}{\textbf{Metamodel-driven Integration:}}
  \\

  \solution{BCM}~(Bruyninckx~\cite{BruyninckxKlotzbucher2013BCM})
  & \xmark
  & \xmark
  & \cmark
  & \cmark
  & \xmark
  & \cmark
  & \cmark
  & \xmark
  & \xmark
  & \xmark
  & \xmark
  & $\blacktriangle$
  & $\blacktriangle$
  & \xmark
  & $\blacktriangle$
  \\

  \arrayrulecolor{gray}\hline\arrayrulecolor{black}

  \multicolumn{16}{l}{\textbf{MBSE-oriented Frameworks:}}
  \\

  \solution{ISE\&PPOOA} (Fernandez~\cite{Hernandez2017Collab})
  & \xmark
  & $\blacktriangle$
  & \cmark
  & \cmark
  & $\blacktriangle$
  & \cmark
  & \cmark
  & $\blacktriangle$
  & \xmark
  & \xmark
  & \xmark
  & \cmark
  & \cmark
  & $\blacktriangle$
  & \cmark
  \\
  \solution{MROS}~(Silva~\citep{Silva2023MROS})
  & \xmark
  & \cmark
  & \cmark
  & \cmark
  & $\blacktriangle$
  & \cmark
  & \cmark
  & $\blacktriangle$
  & \xmark
  & \xmark
  & \xmark
  & \cmark
  & \cmark
  & $\blacktriangle$
  & \cmark
  \\

  \solution{PDD}~(Brambilla~\cite{Brambilla2015PDD})
  & \xmark
  & \xmark
  & \cmark
  & $\blacktriangle$
  & \cmark
  & \cmark
  & \cmark
  & \cmark
  & \xmark
  & \xmark
  & \xmark
  & \xmark
  & \xmark
  & \xmark
  & \xmark
  \\
  \textit{\textbf{Our:}}~\solution{MeROS}~+~V-model (Winiarski~\cite{Winiarski2023MeROS})
  & \xmark
  & \cmark
  & \cmark
  & \cmark
  & \cmark
  & \cmark
  & \cmark
  & \cmark
  & $\blacktriangle$
  & $\blacktriangle$
  & $\blacktriangle$
  & \cmark
  & \cmark
  & \cmark
  & \cmark
  \\
  \bottomrule
  \end{tabular}

  }

  \vspace{2pt}
  \begin{minipage}{\textwidth}
  \footnotesize
  \raggedright
    Marks: \newline
    (\cmark)---supported; \newline
    (\xmark)---not supported; \newline
    ($\blacktriangle$)---partial or informal support (\textit{e.g., missing support in sim or real deployment, or missing executables validation}).
  \end{minipage}

\end{table*}

}

\paragraph{Structural Coverage and Intended Usage}
  \Tab{tab:framework-comparison} summarises structural characteristics captured by the proposed V-model, including traceable abstraction layers, compositional modelling, and behavioural mapping. The model accommodates both top-down architectural planning and reconstruction of legacy systems. These properties support integration across a range of engineering workflows and lifecycle phases. This articulation emphasises validation of the V-model's structure within a mid-phase system context, where architecture and behaviour are defined and embodied.

\paragraph{Framing Key Dimensions of MBSE Maturity}
  Beyond structural clarity, the \solution{MeROS} ecosystem, now backed by V-model integration, offers a framework to address deeper challenges that define the long-term applicability of MBSE in robotics. We identify three foundational dimensions: (1) alignment between \textit{simulation and runtime} modelling (including digital twin capabilities), (2)~\textit{physical grounding} of behaviours in embodied environments, and (3) \textit{compatibility with assurance} processes for certification. These elements define critical readiness checkpoints for any framework aiming to support safety-sensitive or adaptive robotic applications. The following sections examine how \solution{MeROS} engages with these dimensions across the development lifecycle.

\subsection{Simulation-First Development and the Role of Continuous modelling}
\label{sec:simulation-first-development}

  Simulation has become a central asset in modern robotics development workflows. Aligned with the structure of the V-model, simulation enables parallel progress in behaviour specification and validation, decoupled from hardware realisation. This supports early-stage verification and sustains runtime co-evolution via monitoring and iterative updates. Prior work in this research lineage has already demonstrated compatibility between SysML-based architectures and co-simulation infrastructures. The \solution{EARL} modelling approach \cite{Winiarski2020EARL} and subsequent \solution{MeROS} developments \cite{Winiarski2023MeROS} have shown how executable models and system-level abstractions can be integrated into simulation loops. Recent extensions to this methodology have investigated the structural and runtime consistency between simulated models and physical embodiments \cite{Dudek2025SpSysML}, crafting the blueprint for digital twin alignment. In this context, \solution{MeROS} does not implement a digital twin system per se, but its support for co-simulation and architectural traceability allows for digital-twin-like workflows. \model{HeROS}, serving as a minimal but complete physical platform, enables grounded verification of these model-derived assumptions.  This work focuses on how this setup enables lifecycle-wise design steps to align within a \solution{MeROS}-tailored V-model.

\subsection{Certification Readiness of the Lifecycle}
\label{sec:lifecycle-traceability}

  The \solution{MeROS} seeks to refine the \solution{ROS} habits, conscious of the community and industry needs. Building on MBSE principles, \solution{MeROS} provides top-down design and bottom-up validation routines that are important for safety-critical contexts and often missing in ROS runtime-oriented toolchains. Examples like \solution{RTMSafety} \cite{Han2012OPRoS} highlight challenges in achieving certification readiness for open-source platforms. Recently, the work \cite{Winiarski2025ConceptsRico} brought up this question again regarding the potential of open service robotics advancements applications in the specific healthcare context, underlining the importance of \solution{MeROS}'s structured, lifecycle-oriented approach.

  Such fields as automotive and aerospace dictate the trends in lifecycle compliance, and robotics is more of an adopter of best practices at this stage. While \solution{MeROS} does not replicate advanced domain-specific certification features, it furnishes a modelling and verification scaffold compatible with their core expectations, helping systems remain auditable and verification-ready as they scale. The V-model is a particular example of a cross-domain trusted approach, recurring in diverse regulations.

\subsection{From Simulation to Physical Grounding}
\label{sec:sim-2-real}

  Model-based development and simulation-first strategies bring validation earlier in the lifecycle as a source for architectural refinements, but they cannot fully replace actual physical grounding. As R.~Brooks famously noted, ``The world is its own best model'' \cite{Brooks1990Elephants}, reminding, that real-world embodiment is the only reliable test of behaviour validity.

  \solution{MeROS} facilitates system modelling and logic validation at stages prior to deployment. Still, these assumptions are stress-tested only under real-world dynamics via integration with embodied platforms like \model{HeROS} \cite{Winiarski20204HeROS}. Such controlled installations allow for feasibility exploration for the diverse system \solution{MeROS} specifications, enabling versatile validation scenarios with real sensorimotor feedback loops. Details of the system's assessment with similar methodology had been explored recently by Dudek et al. \cite{Dudek2025SpSysML}, who proposed SysML model-derived metrics to quantify fidelity between simulated and embodied behaviours.

  In addition to technical validation, this process supports the epistemic assessment, confirming whether abstractions remain coherent in embodied settings. \solution{MeROS}-tailored V-model develops such a framework even further by providing a structural modelling flow to support the consistency across the abstraction layers. In this way, \solution{MeROS} and \model{HeROS} enable a continuous loop of design, simulation, and grounded evaluation, forming a SysML-documented lifecycle: from abstract models to physical embodiment.

\subsection{Bridging Prior Work and Present Scope}
\label{sec:bridging-prior-work}

  The current formulation unifies two established directions: the metamodelling principles defined in \solution{MeROS} and the embodi\-ment-focused validation enabled by the \model{HeROS} platform \cite{Winiarski20204HeROS}. While this paper centres on the integrated V-model alignment, prior research has already addressed complementary aspects, such as behavioural synchronisation in EARL \cite{Winiarski2020EARL} and structural consistency metrics for simulation vs. physical execution in \solution{SPSysML} \cite{Dudek2025SpSysML}. Certification traceability, too, has been explored through architecture-level annotations in earlier \solution{MeROS} iterations.
  Rather than redefining MBSE, \solution{MeROS} builds on this body of work to offer a robotics-tailored scaffolding that connects modelling, simulation, embodiment, and assurance. The V-model presented here inherits and exposes interfaces for these established assets, enabling systematic integration without demanding conceptual reinvention.
  We summarise the methodology's scope, contributions, and intended trajectory.

\section{Conclusions}
\label{sec:conclusions}

  This work responds to a persistent gap in robotics engineering: the limited availability of structured, lifecycle-aware development practices within the widely adopted \solution{ROS~2} ecosystem. Rather than proposing a universal methodology, we demonstrate how a \solution{MeROS}-aligned adaptation of the V-model can introduce traceability, standards-conscious modelling, and reuse-friendly structure, guiding development from early concept through simulation, integration, and validation.

  Through a representative application on the \model{HeROS} platform, we have shown that this method applies even to heterogeneous, multi-robot systems where embodied feedback plays a key role. By bringing together architectural modelling, real-world embodiment, and certification-aware workflows, our approach doesn't just support core MBSE practices like modular decomposition, co-simulation, and traceable verification---it also makes room for something often overlooked: validating whether our abstractions actually hold up when challenged with the real-world dynamics.

  While \solution{MeROS} and its methodological framing remain under active refinement, the present results confirm that this combination can serve as a practical scaffold for projects facing both engineering complexity and assurance demands. It establishes a continuity between abstract system design and grounded robotic behaviour, informed by simulation, verified in embodiment, and documented for structured development lifecycles.

  This work is not offered as a complete solution, but as a useful and adaptable starting point. We invite reuse, feedback, and critical reflection---whether to build upon our assumptions or challenge them for the betterment of open robotics practice.

\section*{CRediT authorship contribution statement}

\insertcreditsstatement

\section*{Declaration of competing interest}
The authors declare that they have no known competing financial interests or personal relationships that could have appeared to influence the work reported in this paper.

%	\section*{Acknowledgment}
%	The research was funded by the Centre for Priority Research Area Artificial Intelligence and Robotics of Warsaw University of Technology within the Excellence Initiative: Research University (IDUB) programme.

\section*{Data and code availability}
No proprietary datasets were generated or analysed. The MeROS metamodel and model fragments, together with links to supporting modules, are available at:
\url{https://github.com/twiniars/MeROS} and \url{https://github.com/GroupOfRobots/}.

\section*{Supplementary material}
Demonstration videos are available online; links are provided at the relevant points in the manuscript.

\section*{Declaration of Generative AI and AI-assisted technologies in the writing process}
During the preparation of this work, some parts of the language editing were assisted using generative AI tools, including OpenAI's ChatGPT. After using this tool/service, the author(s) reviewed and edited the content as needed and take(s) full responsibility for the content of the publication.

\pagebreak

\balance
\bibliographystyle{elsarticle-num}

\bibliography{MeROS_VModel_HeROS-RAS-Biblio}

\pagebreak
% MeROS_VModel_HeROS-RAS-Preprint-Submission.tex

\section*{Author Biographies}
{

\vspace{1em}

\newcommand{\imH}{1.25in}
\newcommand{\imW}{1in}
\newcommand{\wrapPad}{25mm}
\newcommand{\vSpacing}{10pt}

\begin{wrapfigure}{l}{\wrapPad}
    \includegraphics[
      width=\imW, height=\imH, clip, keepaspectratio]{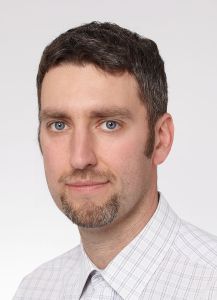}
  \end{wrapfigure}
\par
\textbf{Tomasz Winiarski} IEEE, INCOSE Member, M.Sc./Eng. (2002), PhD (2009) in control and robotics, from Warsaw University of Technology (WUT), assistant professor of WUT. He is a~member of the Robotics Group as the head of the Robotics Laboratory in the Institute of Control and Computation Engineering (ICCE), Faculty of Electronics and Information Technology (FEIT). He is working on the modelling and development procedures of robots in terms of Systems Engineering. The research targets service and social robots as well as didactic robotic platforms. His personal experience concerns the development and modelling of robotic frameworks, manipulator position---force and impedance control, Platform Independent Metamodels (PIM) EARL, SPSysML and Platform Specific Metamodels (PSM) MeROS.
\par

\vspace{\vSpacing}

\begin{wrapfigure}{l}{\wrapPad}
    \includegraphics[width=\imW, height=\imH, clip, keepaspectratio]{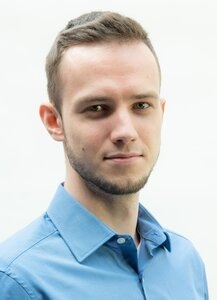}
  \end{wrapfigure}
\par
\textbf{Jan Kaniuka}, M.Sc./Eng. (2024, with honors) in automatic control and robotics from Warsaw University of Technology (WUT), Faculty of Electronics and Information Technology (FEIT). He is a software engineer specializing in robotics, actively participating in the development of cyber-physical systems. His interests focus on systems modelling and embedded systems. He has practical experience in programming mobile robots and manipulators with the Robot Operating System (ROS).
\par

\vspace{\vSpacing}

\begin{wrapfigure}{l}{\wrapPad}
    \includegraphics[width=\imW, height=\imH, clip, keepaspectratio]{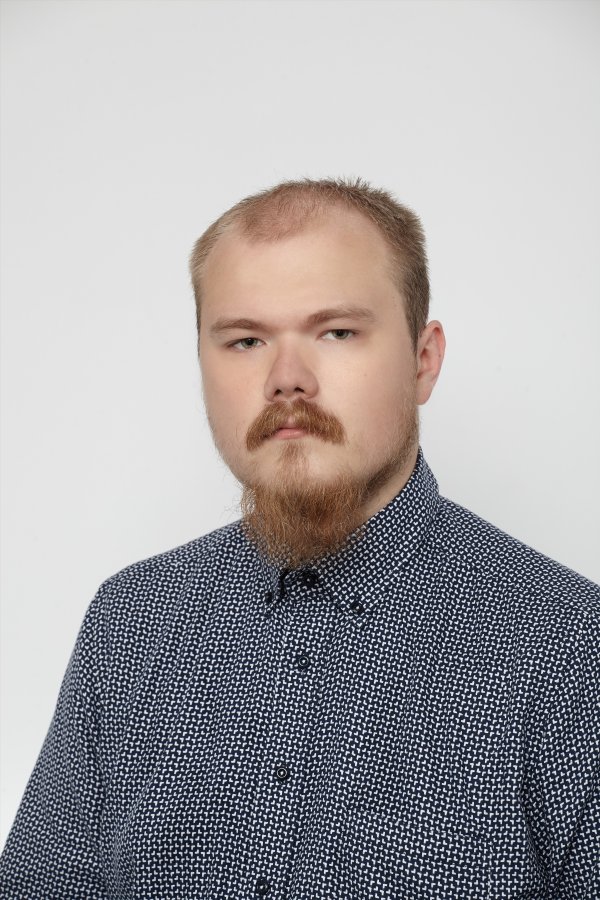}
  \end{wrapfigure}
\par
\textbf{Daniel Giełdowski}, MSc (2021) in automatics control and robotics from Warsaw University of Technology (WUT); Assistant at WUT. Participant of AAL  -- INCARE "Integrated Solution for Innovative Elderly Care", SMIT  -- "Safety-aware Management of robot's Interruptible Tasks in dynamic environments", and LaVA  -- „Laboratory for testing the vulnerability of stationary and mobile IT devices as well as algorithms and software” projects. Focused on artificial intelligence and cyber security of robotic algorithms.
\par

\vspace{\vSpacing}

\begin{wrapfigure}{l}{\wrapPad}
    \includegraphics[
      width=\imW, height=\imH, clip, keepaspectratio]{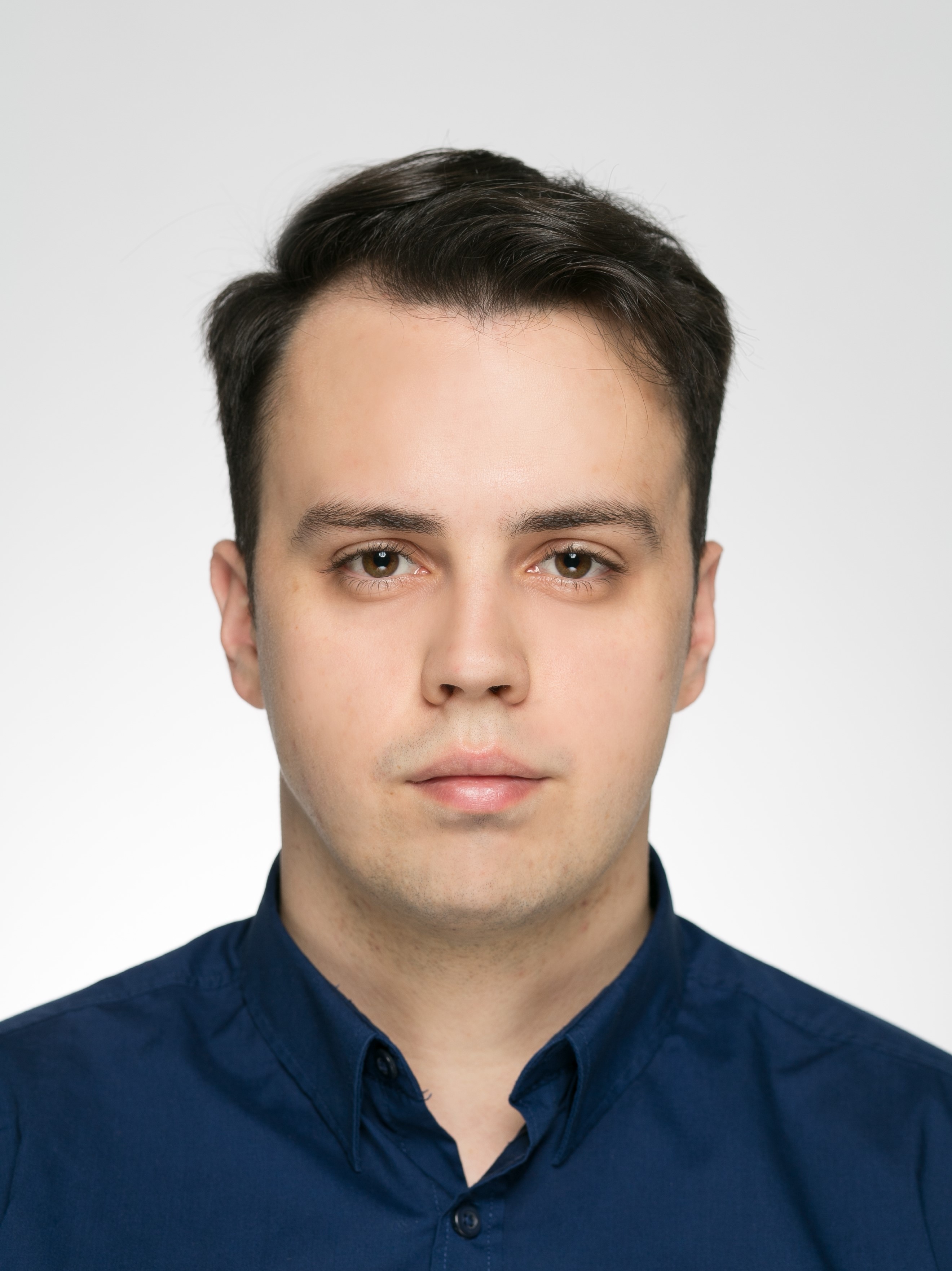}
\end{wrapfigure}
\par
\textbf{Jakub Ostrysz},  M.Sc./Eng. (2025, with honors) in automatic control and robotics from Warsaw University of Technology (WUT), Faculty of Electronics and Information Technology (FEIT). His main research interests include multi-robot systems and robot swarms. He is a software engineer focusing on the design and development of advanced navigation systems for mobile robots. His professional experience includes the design and implementation of navigation systems for robot swarms and the development of autonomous mobile robots (AMR).
\par

\vspace{\vSpacing}

\begin{wrapfigure}{l}{\wrapPad}
    \includegraphics[width=\imW, height=\imH, clip, keepaspectratio, trim=25 0 25 0]{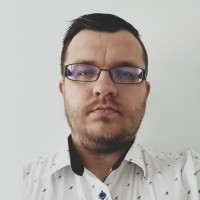}
\end{wrapfigure}
\par
\textbf{Krystian Radlak}, B.Sc./M.Sc. bioinformatics (2010, 2011), and B.Sc. in mathematics (2012), PhD in automatic control and robotics (2017) from the the Silesian University of Technology, assistant professor at WUT. He also work as a senior consultant at UL Solutions. His current research interests include low-level image processing, computer vision and application functional safety standards for AI/ ML safety critical systems.
\par

\vspace{\vSpacing}

\begin{wrapfigure}{l}{\wrapPad}
    \includegraphics[
      width=\imW, height=\imH, clip, keepaspectratio, trim=20 0 20 0]{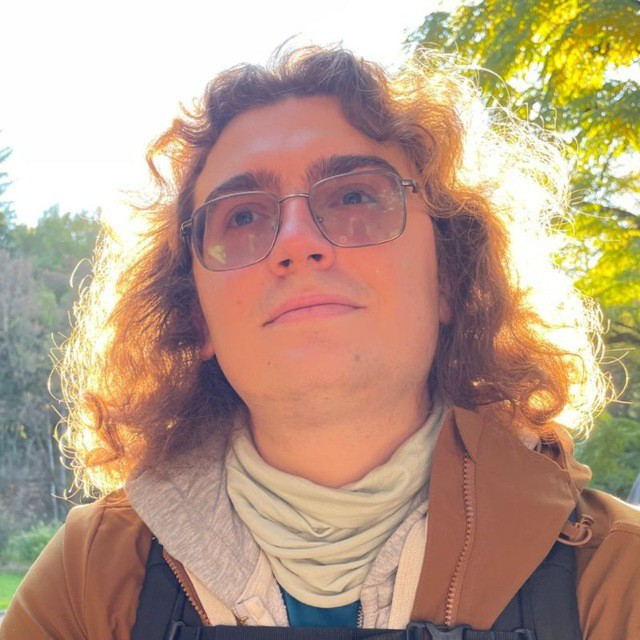}
\end{wrapfigure}
\par
\textbf{Dmytro Kushnir}, B.Sc. and M.Sc. in Law (2014, 2015). B.Sc. in Computer Science (2019, summa cum laude) from the Ukrainian Catholic University (UCU). Currently an Assistant Professor and Ph.D. candidate at UCU. His experience includes industrial robotics and low-level image processing. His research focuses on outdoor navigation and model-based systems engineering. He joined the project through his involvement in UGV system development and interest in applying MeROS and V-model-based methodologies for autonomous UGV.
\par

}

\end{document}